\documentclass[conference]{IEEEtran}

\usepackage{times}
\usepackage{epsfig}
\usepackage{graphicx}
\usepackage{amsmath}
\usepackage{amssymb}

\usepackage{enumerate}
\usepackage{multirow}
\usepackage{tabularx}
\usepackage{nicefrac}
\usepackage{booktabs}
\usepackage{enumitem}

\usepackage{makecell}
\usepackage{xspace}
\usepackage{graphbox}
\usepackage{pifont}
\usepackage{enumitem}
\usepackage[dvipsnames]{xcolor}
\usepackage{colortbl}
\usepackage{multirow}
\usepackage{array}
\usepackage{caption}
\usepackage[linesnumbered,ruled]{algorithm2e}
\usepackage{wrapfig}
\usepackage{glossaries}
\glsdisablehyper
\usepackage{multirow}
\usepackage{xspace}
\usepackage{tikz}
\usepackage{bbm}
\usepackage{soul}

\usepackage[pagebackref=true,breaklinks=true,letterpaper=true,colorlinks,bookmarks=false]{hyperref}
\usepackage{xurl}

\graphicspath{{figures}}

\def\R{\mathbb{R}}

\newcommand{\norm}[1]{\left\|#1\right\|}

\def\L{\mathcal{L}}

\def\maxop{\mathop{\rm max}\limits}

\newcolumntype{C}[1]{>{\centering\arraybackslash}p{#1}}
\newcolumntype{L}[1]{>{\raggedright\arraybackslash}p{#1}}
\newcolumntype{R}[1]{>{\raggedleft\arraybackslash}p{#1}}

\definecolor{Gray}{gray}{0.85}
\definecolor{BlueGray}{rgb}{0.9, 0.9, 1}
\definecolor{LightCyan}{rgb}{0.88,1,1}
\definecolor{lightgrey}{rgb}{0.9, 0.9, 0.9}

\newcommand{\hide}[1]{}

\newlength\newl
\newlength\colwidth
\newlength\tabrowsep
\newlength\rowheight

\newacronym{bpda}{BPDA}{Backward Pass Differentiable Approximation}
\newacronym{eot}{EoT}{Expectation over Transformation}
\newacronym{pgd}{PGD}{Projected Gradient Descent}
\newacronym{awp}{AWP}{Adversarial Weight Perturbation}
\newacronym{ebm}{EBM}{Energy-Based Model}
\newacronym{dsm}{DSM}{Denoising Score-Matching}
\newacronym{adp}{ADP}{Adaptive Denoising Purification}
\newacronym{apgd}{APGD}{AutoPGD}
\newacronym{fgsm}{FGSM}{Fast Gradient Sign Method}
\newacronym{pca}{PCA}{Principal Component Analysis}
\newacronym{mcmc}{MCMC}{Markov chain Monte Carlo}
\newacronym{hd}{HD}{Hedge Defense}
\newacronym{dlr}{DLR}{Difference of Logits Ratio}
\newacronym{ode}{ODE}{Ordinary Differential Equation}
\newacronym{cw}{CW}{Carlini-Wagner}
\newacronym{atld}{ATLD}{Adversarial Training with Latent Distribution}
\newacronym{imf}{IMF}{Inference with Manifold Transformation}

\newcommand{\sam}{SAM\xspace}
\newcommand{\mobilesam}{MobileSAM\xspace}
\newcommand{\seem}{SEEM\xspace}

\newcommand{\ade}{\textsc{Ade20K}\xspace}
\newcommand{\saoneb}{\textsc{SA-1B}\xspace}
\newcommand{\coco}{\textsc{Coco}\xspace}

\usepackage{amsmath,amsfonts,bm}

\def\1{\bm{1}}

\def\vdelta{{\bm{\delta}}}

\def\vm{{\bm{m}}}

\def\vx{{\bm{x}}}

\DeclareMathAlphabet{\mathsfit}{\encodingdefault}{\sfdefault}{m}{sl}
\SetMathAlphabet{\mathsfit}{bold}{\encodingdefault}{\sfdefault}{bx}{n}

\def\BibTeX{{\rm B\kern-.05em{\sc i\kern-.025em b}\kern-.08em T\kern-.1667em\lower.7ex\hbox{E}\kern-.125emX}}

\begin{document}

\title{
Segment (Almost) Nothing: Prompt-Agnostic Adversarial Attacks on Segmentation Models
}

\author{
\IEEEauthorblockN{Francesco Croce}
\IEEEauthorblockA{
\textit{University of T{\"u}bingen, T{\"u}bingen AI Center}
}
\and
\IEEEauthorblockN{Matthias Hein}
\IEEEauthorblockA{\textit{University of T{\"u}bingen, T{\"u}bingen AI Center}}

}

\maketitle

\begin{abstract}
   General purpose segmentation models are able to generate (semantic) segmentation masks from a variety of prompts, including visual (points, boxed, etc.) and textual (object names) ones. In particular, input images are pre-processed by an image encoder to obtain embedding vectors which are later used for mask predictions.
   Existing adversarial attacks target the end-to-end tasks, i.e. aim at altering the segmentation mask predicted for a specific image-prompt pair. However, this requires running an individual attack for each new prompt for the same image.
   We propose instead  to generate prompt-agnostic adversarial attacks by maximizing the $\ell_2$-distance, in the latent space, between the embedding of the original and perturbed images. Since the encoding process only depends on the image, 
   distorted image representations will cause perturbations in the segmentation masks for a variety of prompts.
   We show that even imperceptible $\ell_\infty$-bounded perturbations of radius $\epsilon=1/255$ are often sufficient to drastically modify the masks predicted with point, box and text prompts by recently proposed foundation models for segmentation. 
   Moreover, we explore the possibility of creating universal, i.e. non image-specific, attacks which can be readily applied to any input without further computational cost. 
   
\end{abstract}


\section{Introduction}

Foundation models, that is large pre-trained models which can be easily adapted to downstream tasks, have been recently proposed in a variety of domains \cite{radford2021learning, touvron2023llama}.
In the context of segmentation, general purpose models like \sam \cite{kirillov2023segany} and \seem \cite{zou2023segment} are able to segment objects given visual, text or audio prompts, or even provide a (semantic) segmentation map for an entire image without any specific prompts.
These models exhibit strong generalization performance to unseen datasets, which makes them well-suited for being readily deployed in a multitude of practical applications. 
This emphasizes the relevance of understanding their robustness to adversarial attacks, since their potential vulnerabilities might threaten the safety of systems integrating these models.
However, this aspect has been so far only partially studied: in fact, previous works \cite{zhang2023attack, huang2023robustness, qiao2023robustness} analyze the robustness of \sam when an end-to-end task is attacked, which means that one generates an adversarial perturbation to alter the predicted mask for a given image-prompt pair.
While this yield strong task-specific attacks, it requires running the attack for each new prompt (and image) independently, which might be expensive, and the attack circumvented by using different prompts.

\begin{figure*}[t] \centering
\small
\includegraphics[width=1.9\columnwidth]{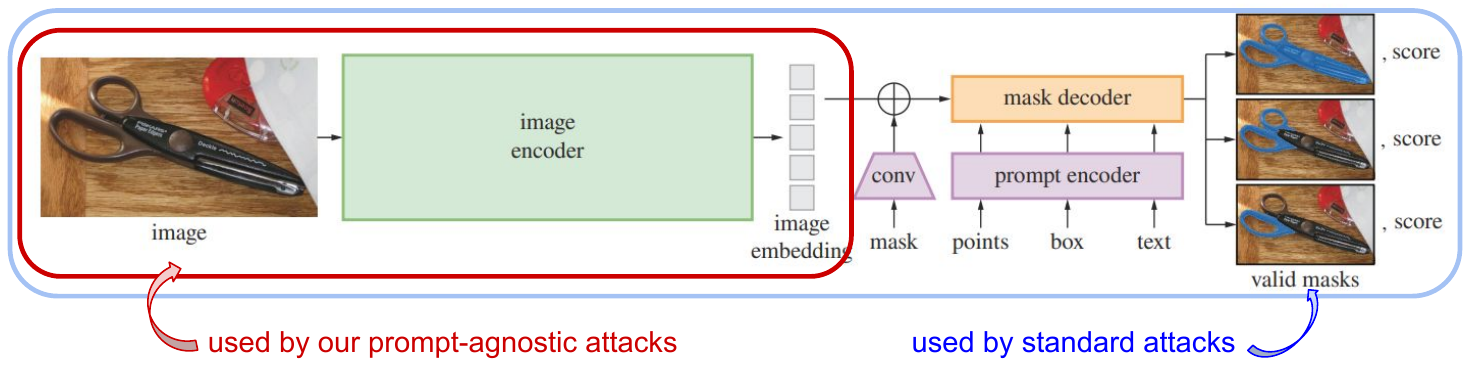}
\caption{\textbf{Visualization of the architecture of \sam and our proposed method.} We show the structure of the architecture of \sam (image taken from \cite{kirillov2023segany}) and different attack approaches. While standard attacks consider all components of the segmentation model and the predicted masks to generate their perturbations, we propose to only distort the embedding provided by the image encoder. In this way, the resulting perturbations are not specific to the prompt used for generating the attacks.}
\label{fig:vis_sam}
\end{figure*}

In this work, we exploit the fact that foundation segmentation models process an input image via an encoder which does not take into account the prompts defining the segmentation tasks. In this way, the image embedding is computed once, as it is typically computationally intense, and then re-used to generate the segmentation masks corresponding to multiple prompts.
We propose to generate prompt-agnostic adversarial attacks by modifying an input image to distort the embedding produced by the image encoder, i.e. maximize the $\ell_2$-distance between the representations, in latent space, of clean and adversarial image (see visualization in Fig.~\ref{fig:vis_sam}).
The intuition behind this approach is that, if the embedding of the adversarial image is sufficiently altered compared to the original one, it will not be useful to obtain precise segmentation masks regardless of the type and instance of prompts received.
Moreover, this allows us to compute a single attack which effectively degrades the performance of the model independently of which task it is used for.

With this approach we show that small $\ell_\infty$-norm bounded perturbations, which can be optimized with standard techniques for adversarial attacks like projected gradient descent \cite{MadEtAl2018}, are able to significantly degrade the performance of \sam and \seem on a variety of tasks, while introducing imperceptible changes to the original image (Sec.~\ref{sec:white-box_sam} and Sec.~\ref{sec:white-box_seem}).
Moreover, we adapt our algorithm to generate universal adversarial perturbations (Sec.~\ref{sec:universal_attacks_sam}), that is a single perturbation is generated leveraging a limited number of training images and then can be applied to any new unseen image without additional cost.
Finally, in Sec.~\ref{sec:multicrop_generator} we provide a version of our attacks to counter the use of a more sophisticated and expensive configuration of the mask generator of \sam.

\section{Related Work} \label{sec:related_work}

\textbf{Robustness of \sam.}
\cite{zhang2023attack} first studied the adversarial robustness of \sam: they generate $\ell_\infty$-bounded perturbations with either FGSM \cite{GooShlSze2015} or PGD \cite{MadEtAl2018} (with 10 steps) to remove or manipulate the predicted mask for a given pair of image and prompt. 
Similarly, \cite{huang2023robustness} use 10 steps of various algorithms (BIM \cite{KurGooBen2016a}, PGD, SegPGD \cite{gu2022segpgd}) for $\ell_\infty$-bounded attacks to alter the original predicted mask for a given prompt by maximizing the training loss of \sam, i.e. a combination of Focal \cite{lin2017focal} and Dice \cite{sudre2017generalised} loss.
\cite{qiao2023robustness} further conducts an evaluation similar to \cite{zhang2023attack}, increasing the number of iterations in PGD to 20.
These works focus on the task of removing or changing the predicted mask for a specific point prompt, and show the vulnerability of \sam to attacks targeting it.
In \cite{zhang2023attack} \emph{cross-prompt} tasks are also introduced, where an attack generated with a source prompt is evaluated with a different target prompt, and show that increasing the number of source prompts (up to 400) to optimize the adversarial perturbations might improve its generalization to unseen prompts. However, only single point prompts are used as targets.
Finally, \cite{huang2023robustness, qiao2023robustness} study the performance of \sam when the input images are changed by various common (non-adversarial) corruptions e.g. those in ImageNet-C \cite{HenDie2019} or style transfer.
\\

\textbf{Adversarial attacks for semantic segmentation.}
Several works have focused on developing adversarial attacks against semantic segmentation models: most research \cite{hendrik2017universal,arnab2018robustness,mopuri2018generalizable,gu2022segpgd,agnihotri2023cospgd,croce2023robust,halmosi2023evaluating} has considered the widely popular $\ell_\infty$-bounded threat model, while some attention has been received by patch attacks \cite{nesti2022evaluating} and unconstrained perturbations \cite{xie2017adversarial, shen2019advspade, kang2020adversarial, rony2022proximal}.
Moreover, some of these works have introduced universal \cite{hendrik2017universal, kang2020adversarial} and data-free \cite{mopuri2018generalizable} attacks against semantic segmentation models.

Related to our approach, \cite{mopuri2018generalizable, halmosi2023evaluating} introduce techniques which involve attacking some internal representation of the target models.
In particular, in 
recent work, 
\cite{halmosi2023evaluating} consider attacking a semantic segmentation model maximizing the cosine similarity between the output of the model backbone for the adversarial image and a random vector (or a target vector for targeted attacks). Additionally they combine such objective function with a loss on the predictions of the entire model (as standard for attacks on semantic segmentation tasks), and solve the resulting optimization problem with Adam \cite{KinEtAl2014} together with a projection operation.
Unlike \cite{halmosi2023evaluating}, we do not rely on a target vector in the embedding space or additional losses including the predicted masks.

In general, distorting internal representations for adversarial attacks has been explored in the literature, even in the context of attacks on image classifiers e.g. by \cite{sabour2016adversarial, inkawhich2019feature}.
However, compared to \cite{halmosi2023evaluating,mopuri2018generalizable}, our work is, to our knowledge, the first one to focus on prompt-agnostic (universal) attacks: by targeting the image encoder of foundation models, we obtain adversarial perturbations which are independent of a specific downstream segmentation task and its associated performance metric.

\section{Prompt-Agnostic Adversarial Attacks}

The architecture of both \sam, \mobilesam \cite{zhang2023faster} (a lightweight version of \sam) and \seem consists of the following elements (see Fig.~\ref{fig:vis_sam} for an illustration): input images, visual and text prompts are processed separately by different encoders, whose outputs are then combined to generate the predicted masks (and possibly classes, when semantic segmentation is supported as in \seem).
We can formalize such models as a function $f$ which, given an image $\vx$ and set of prompts $P$, returns a mask $\vm=f(\vx, P)$ with the predicted segmentation.
Previous works \cite{zhang2023attack, huang2023robustness, qiao2023robustness} have generated adversarial perturbations against \sam by solving 
\begin{align} \begin{split} \maxop_{\vdelta \in \R^{w\times h \times c}} \; &\L\big(f(\vx + \vdelta, P), f(\vx, P)\big) \\
&\textrm{s.th.} \quad \norm{\vdelta}_p \leq \epsilon, \quad \vx + \vdelta \in [0, 1]^{w\times h \times c},
\end{split} \label{eq:prompt-specific-attacks}
\end{align}
for some loss function $\L$, where $\epsilon$ is the largest $\ell_p$-norm allowed for the perturbation. 
This attack aims at changing the prediction of $f$ for specific image-prompt pairs $(\vx, P)$, but need not affect the outcome for a different prompt $P$ as has been shown in \cite{zhang2023attack}.

However, as part of the model $f$, a vision backbone $\phi : \R^{w\times h \times c} \longrightarrow \R^n$ is used as image encoder to extract a feature vector $\phi(\vx)$ for the input image $\vx$ which is at this stage independent of the prompt $P$, see Fig.~\ref{fig:vis_sam}.
Thus we propose to use instead the following attack objective:
\begin{align} \begin{split} \maxop_{\vdelta \in \R^{w\times h \times c}} \; &\norm{\phi(\vx + \vdelta) - \phi(\vx)}_2^2\\
&\textrm{s.th.} \quad \norm{\vdelta}_p \leq \epsilon, \quad \vx + \vdelta \in [0, 1]^{w\times h \times c},
\end{split} \label{eq:prompt-agnostic-attacks}
\end{align}
where, compared to Eq.~\eqref{eq:prompt-specific-attacks}, we remove the dependence on $P$. 
Intuitively, maximally perturbing the features extracted by the image encoder should mislead the downstream segmentation output regardless of the prompt provided by the user.

The optimization problems in Eq.~\eqref{eq:prompt-specific-attacks} and Eq.~\eqref{eq:prompt-agnostic-attacks} can be solved by Projected Gradient Descent (PGD) \cite{MadEtAl2018} and its variants \cite{KurGooBen2016a, croce2020reliable} similar to $\ell_p$-bounded attacks on image classification. 
We use APGD \cite{croce2020reliable} with an automatically adaptive schedule and stepsize which has been shown to outperform standard PGD \cite{MadEtAl2018}. However, APGD can only be used if the objective is fixed as it tracks the objective, e.g. for the stepsize selection. Thus for varying objectives as in universal attacks we use normal PGD.

\begin{figure*}[p] \centering \small
\newl=1.9\columnwidth
\colwidth=.4\columnwidth
\tabcolsep=1.1pt 
\begin{tabular}{*{5}{C{\colwidth}}}
& original & $1/255$ & $2/255$ & $4/255$ \\
\toprule

\multicolumn{5}{c}{
\includegraphics[width=\newl]{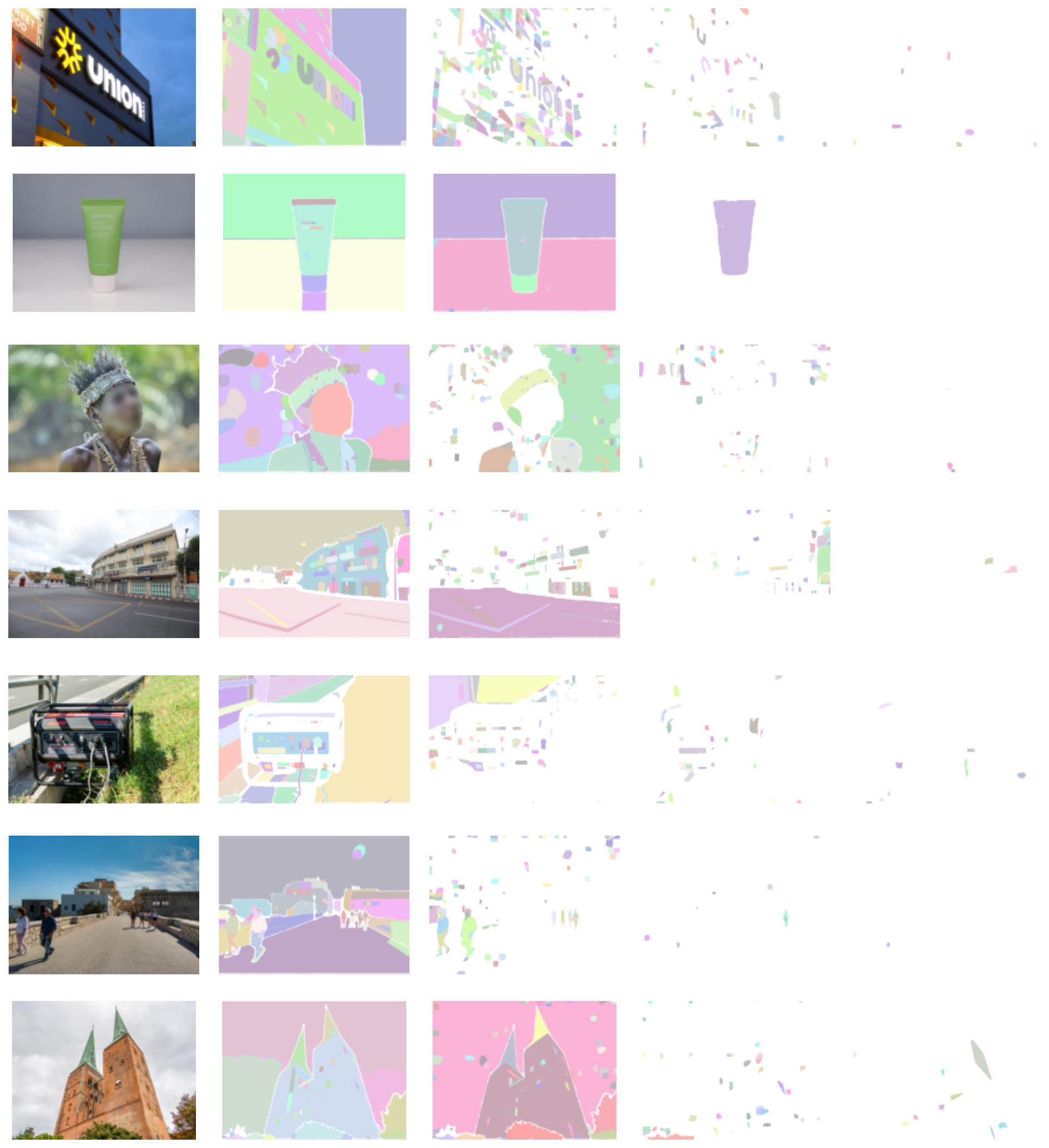}}

\end{tabular}
\caption{\textbf{Segment Everything mode of \sam.} We show the effect of adversarial perturbations with different $\ell_\infty$ bounds $\epsilon=\{1/255, 2/255, 4/255\}$ on the results of Segment Everything mode of \sam. Even small perturbations are sufficient to significantly modify the predicted segmentation masks.}
\label{fig:sam_segment_everything}
\end{figure*}

\section{Experiments}

\subsection{White-box attacks on \sam} \label{sec:white-box_sam}

In the following we test our attack in the white-box scenario (the attacker has complete access to the target model) using the publicly available\footnote{We use the implementation and models provided at \url{https://github.com/facebookresearch/segment-anything}.} checkpoint for the ViT-H \cite{dosovitskiy2020image} backbone, and the default parameters for the mask generators.
If not stated otherwise we use 100 steps of APGD \cite{croce2020reliable} to generate $\ell_\infty$-bounded adversarial perturbations by solving the problem in Eq.~\eqref{eq:prompt-agnostic-attacks}.
Note that this is a relatively small computational budget for adversarial attacks, which typically rely on thousands of iterations, possibly distributed over several random restarts \cite{croce2020reliable}.
We fix such budget since it already provides strong attacks and keeps a relatively small computational cost, given the large architectures used by the target models.
As images we use random samples from the 
a random chunk of the \saoneb dataset \cite{kirillov2023segany}.
\\

\textbf{Segment Everything mode.}
We first consider the mode of \sam in which it tries to segment all existing objects in an image. For this, a grid of point prompts is automatically generated, and their predicted masks are then automatically filtered and deduplicated.\footnote{See e.g. \url{https://github.com/facebookresearch/segment-anything/blob/main/notebooks/automatic_mask_generator_example.ipynb} for more details.}
In Fig.~\ref{fig:sam_segment_everything} we show, for each row, first the original input image and its predicted masks (i.e. no perturbations is added), then the predicted segmentation masks when a perturbation of size $\epsilon$ is applied, with increasing values $\epsilon \in \{1/255, 2/255, 4/255\}$.
We remark that these radii are sufficiently small to introduce perturbations which are not visible in the large majority of case, and in the range of what commonly used in works of adversarial robustness of both classification \cite{debenedetti2022light} and segmentation \cite{croce2023robust}.
One can observe that already the smallest perturbations ($\epsilon=1/255$) are sufficient to significantly alter the model predictions for most cases.
Note that the white areas indicate that no mask is predicted for those pixels.
Some of the largest areas with uniform colors are still correctly segmented, but disappear with larger $\epsilon$ values.
\\

\begin{figure*} \centering \small
\newl=1.02\columnwidth
\tabcolsep=1.1pt 
\begin{tabular}{*{5}{C{17mm}} | *{5}{C{17mm}}}
\multicolumn{5}{c|}{\textbf{seed \#1}} & \multicolumn{5}{c}{\textbf{seed \#2}}\\[2mm]
original & $1/255$ & $2/255$ & $4/255$ & $8/255$ & original & $1/255$ & $2/255$ & $4/255$ & $8/255$ \\
\toprule
\multicolumn{5}{c|}{\includegraphics[align=c, width=\newl]{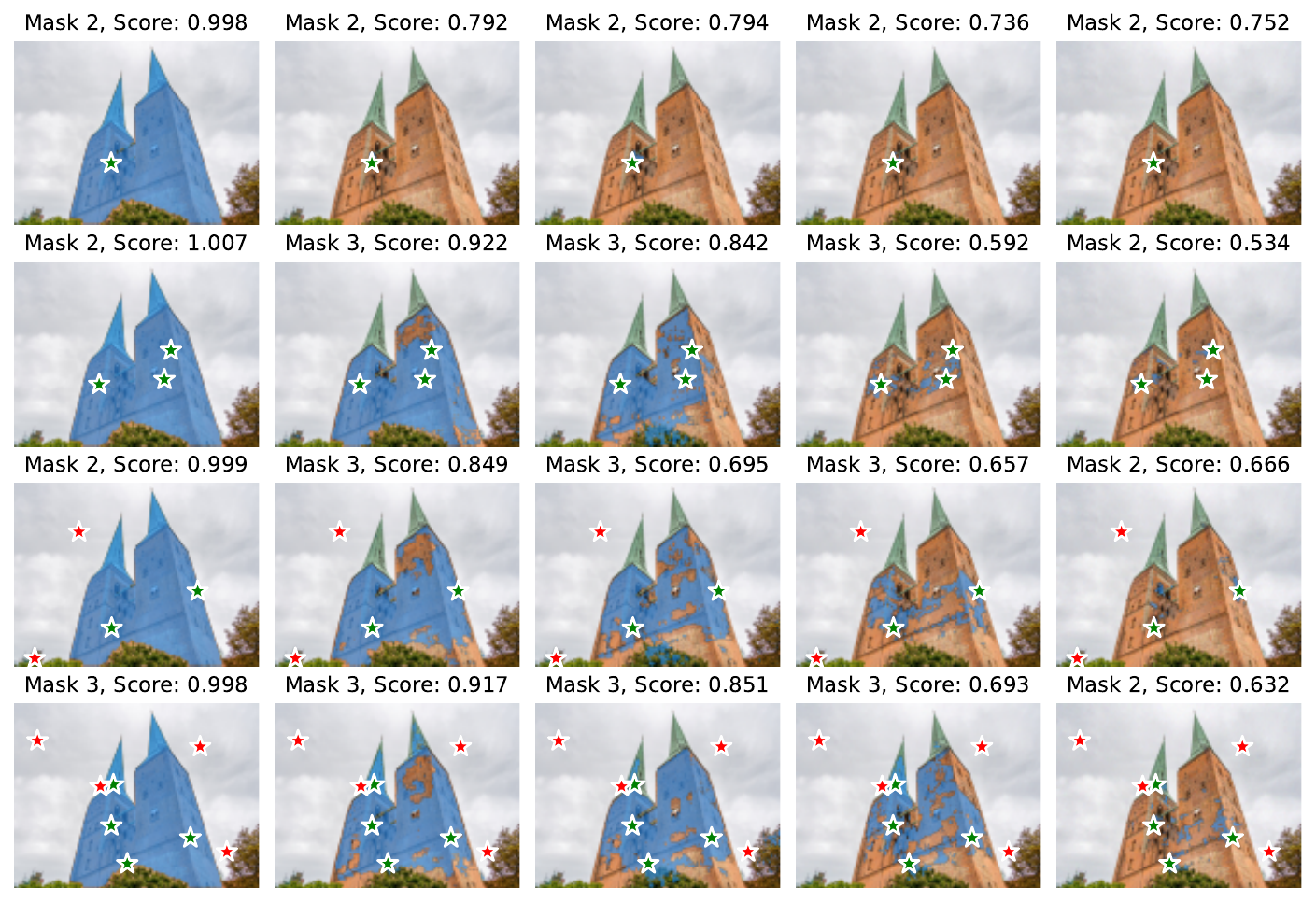}} & \multicolumn{5}{c}{\includegraphics[align=c, width=\newl]{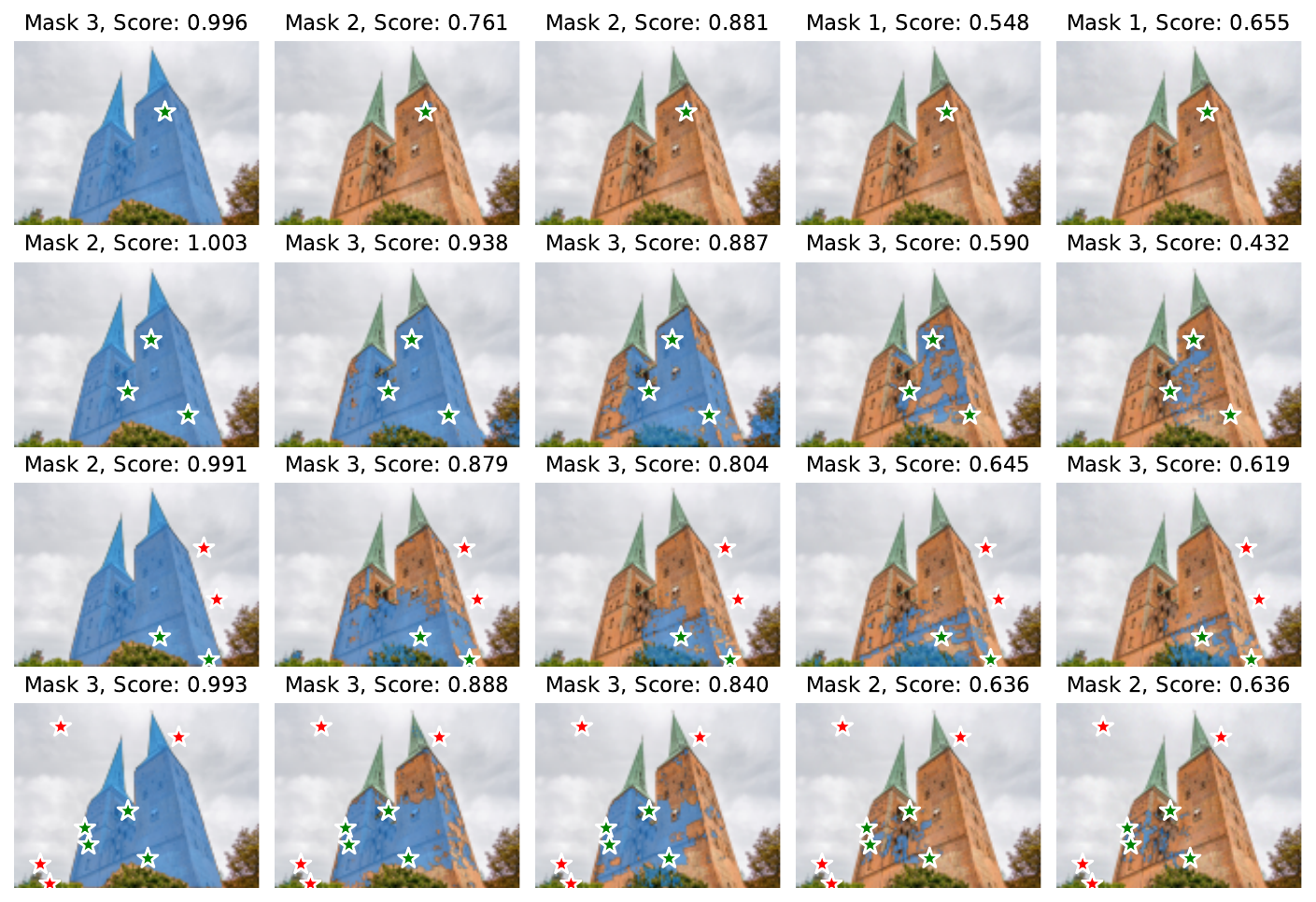}}\\
\midrule
\multicolumn{5}{c|}{\includegraphics[align=c, width=\newl]{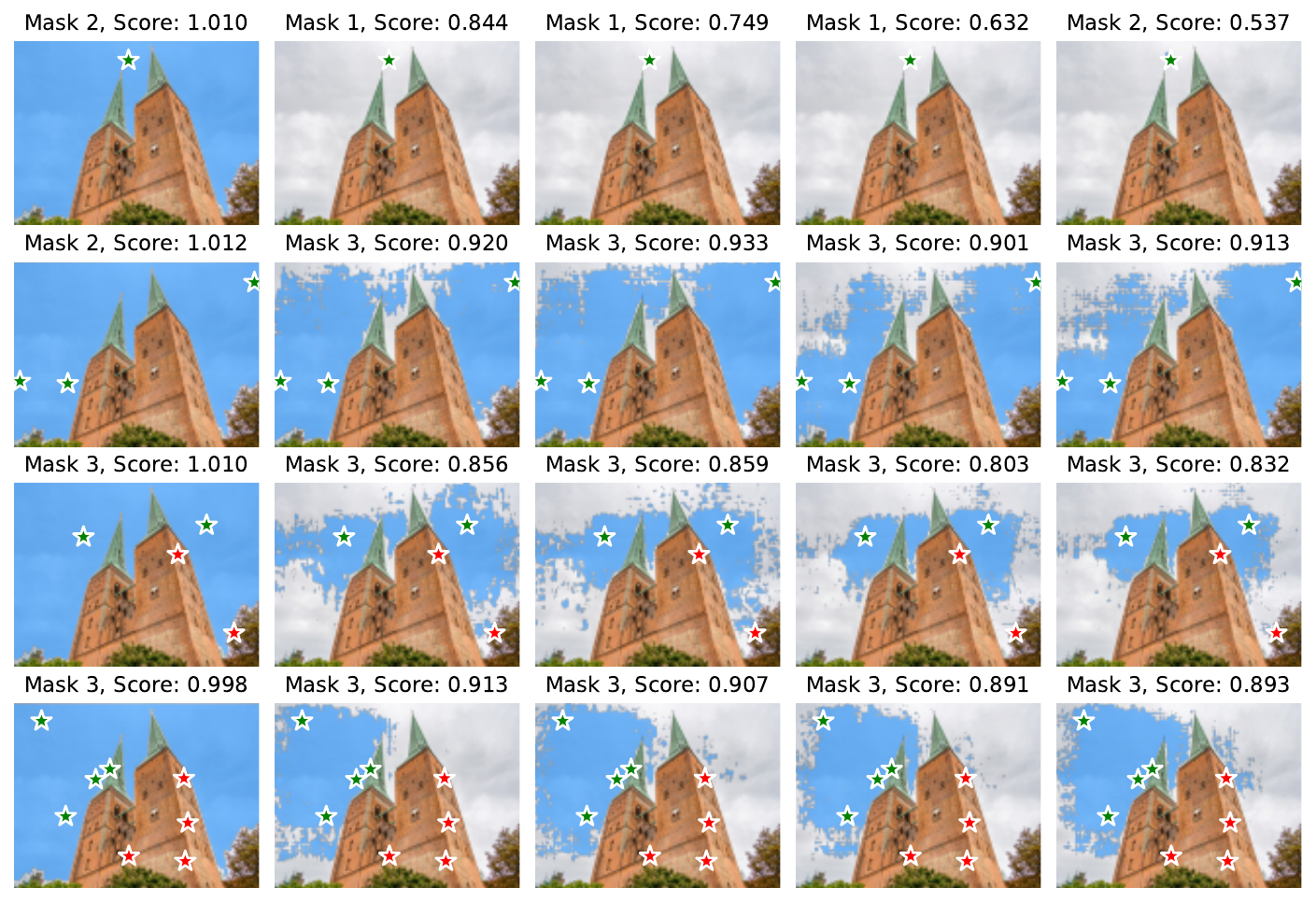}} & \multicolumn{5}{c}{\includegraphics[align=c, width=\newl]{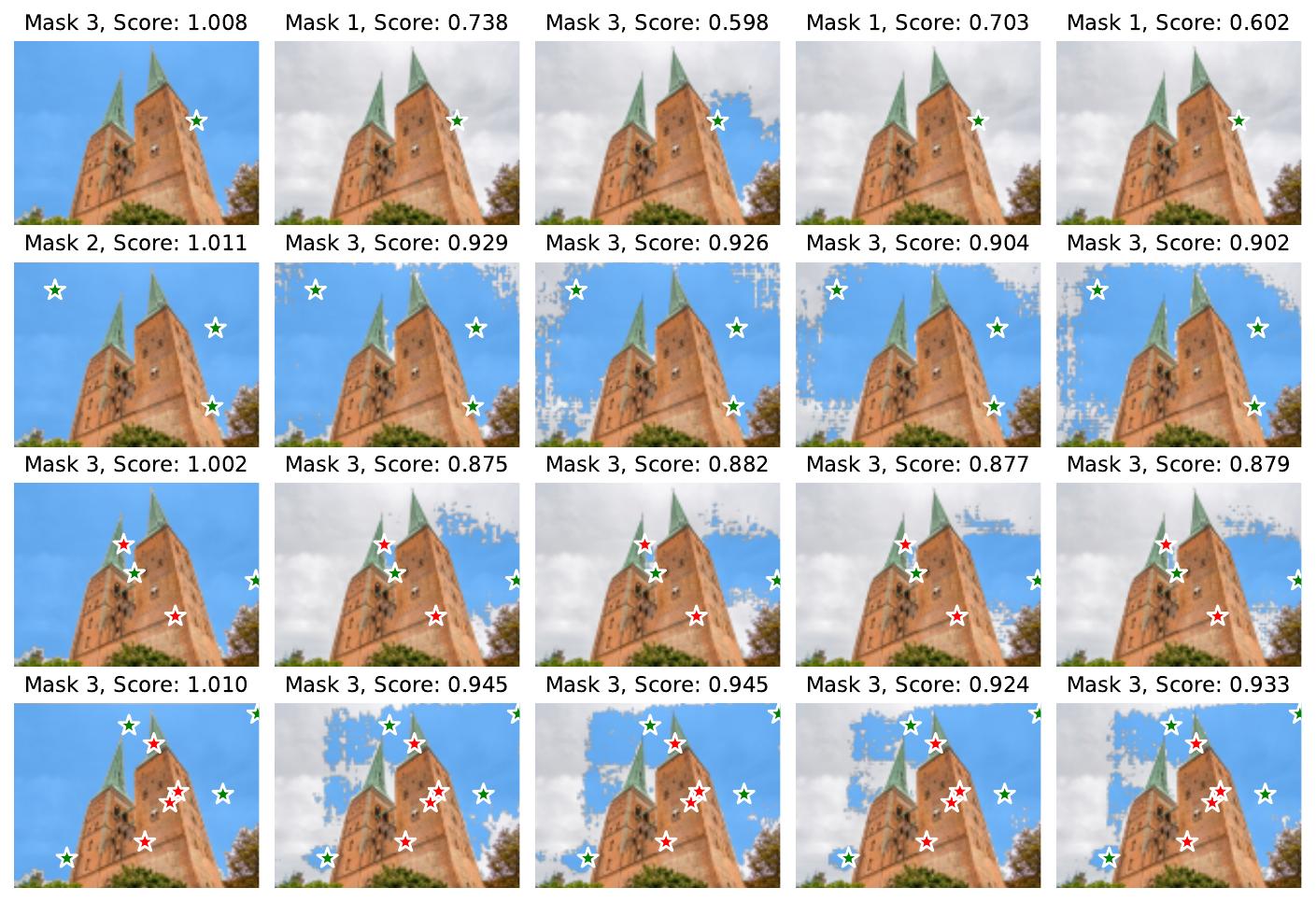}}

\end{tabular}
\caption{\textbf{Multiple point prompts on \sam.} For each mask (different blocks) we vary the number of randomly selected positive (green stars) and negative (red stars) point prompts (different rows), and repeat with two seeds (left and right sides of the panel) which lead to sampling different point prompts. Above each image we report the quality score predicted by the model for the mask (the mask with highest score is selected among the three proposals). For adversarially perturbed images (radii $\epsilon \in \{1/255, 2/255, 4/255, 8/255\}$) the quality of the masks is significantly degraded, even when increasing the number of point prompts.} \label{fig:sam-multi-point-prompts}

\end{figure*}

\begin{figure*} \centering \small
\newl=1.02\columnwidth
\tabcolsep=1.1pt 
\begin{tabular}{*{5}{C{17mm}} | *{5}{C{17mm}}}
\multicolumn{5}{c|}{\textbf{seed \#1}} & \multicolumn{5}{c}{\textbf{seed \#2}}\\[2mm]
original & $1/255$ & $2/255$ & $4/255$ & $8/255$ & original & $1/255$ & $2/255$ & $4/255$ & $8/255$ \\
\toprule

\multicolumn{5}{c|}{\includegraphics[align=c, width=\newl]{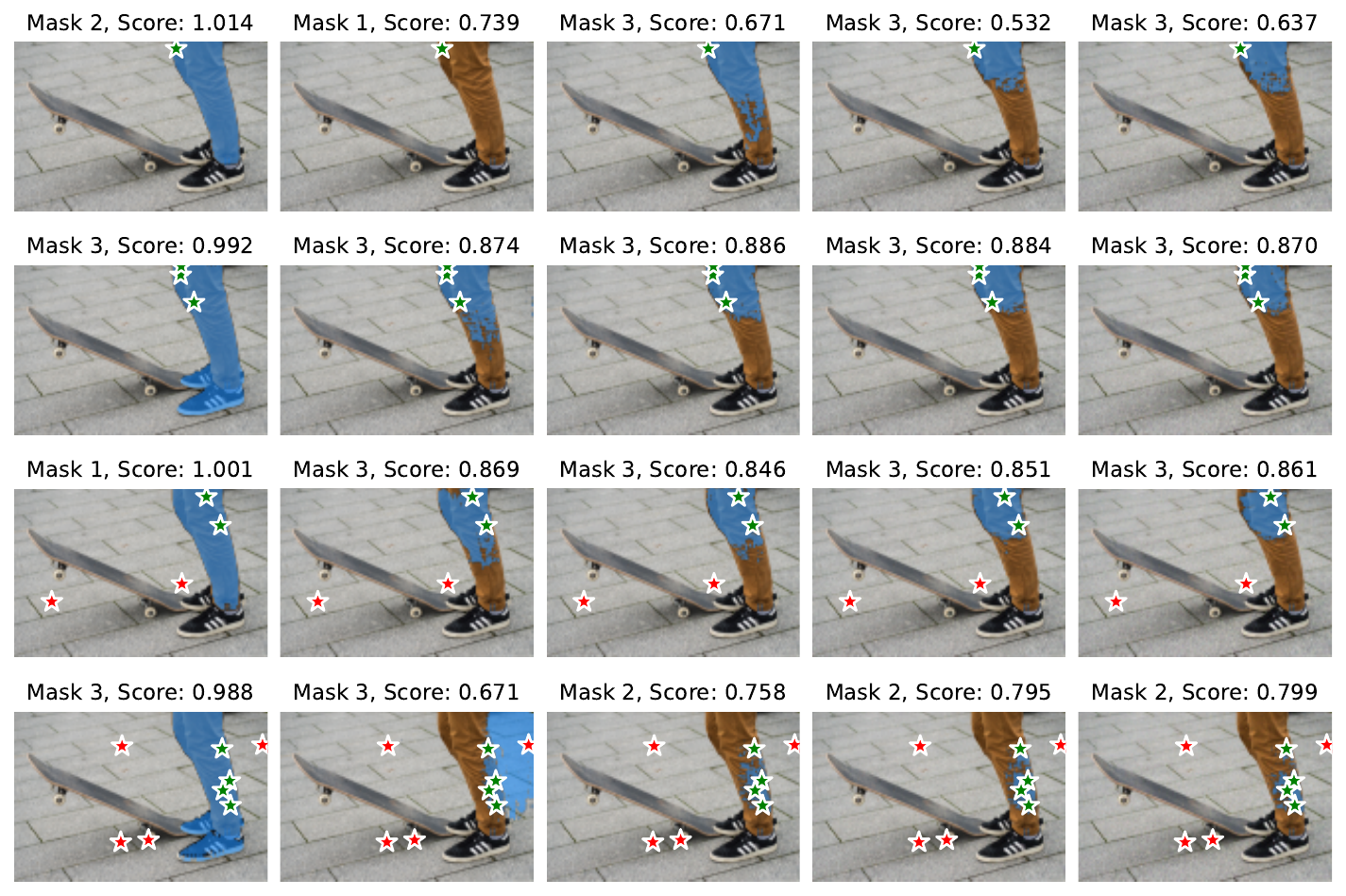}} & \multicolumn{5}{c}{\includegraphics[align=c, width=\newl]{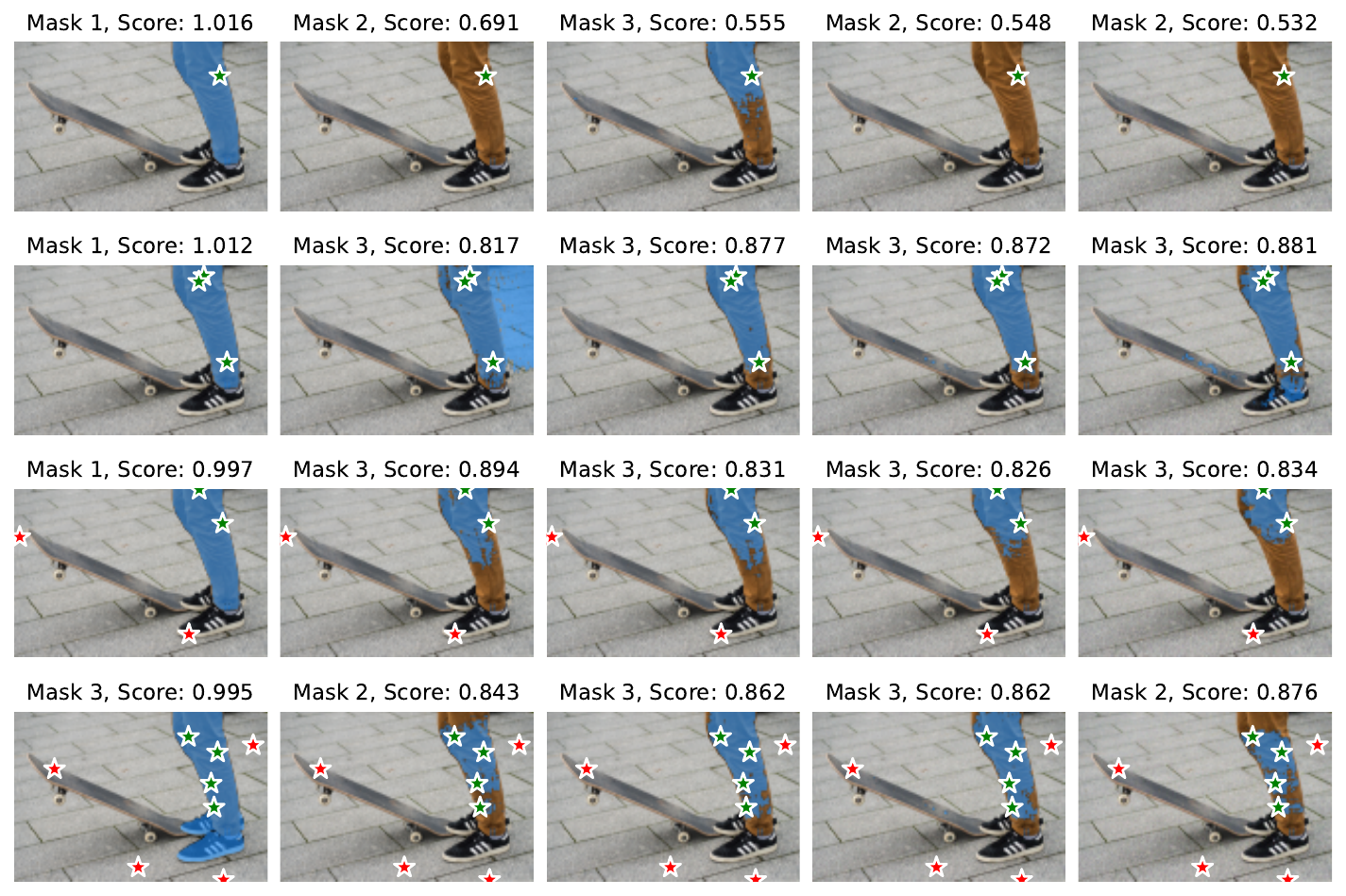}} \\
\midrule
\multicolumn{5}{c|}{\includegraphics[align=c, width=\newl]{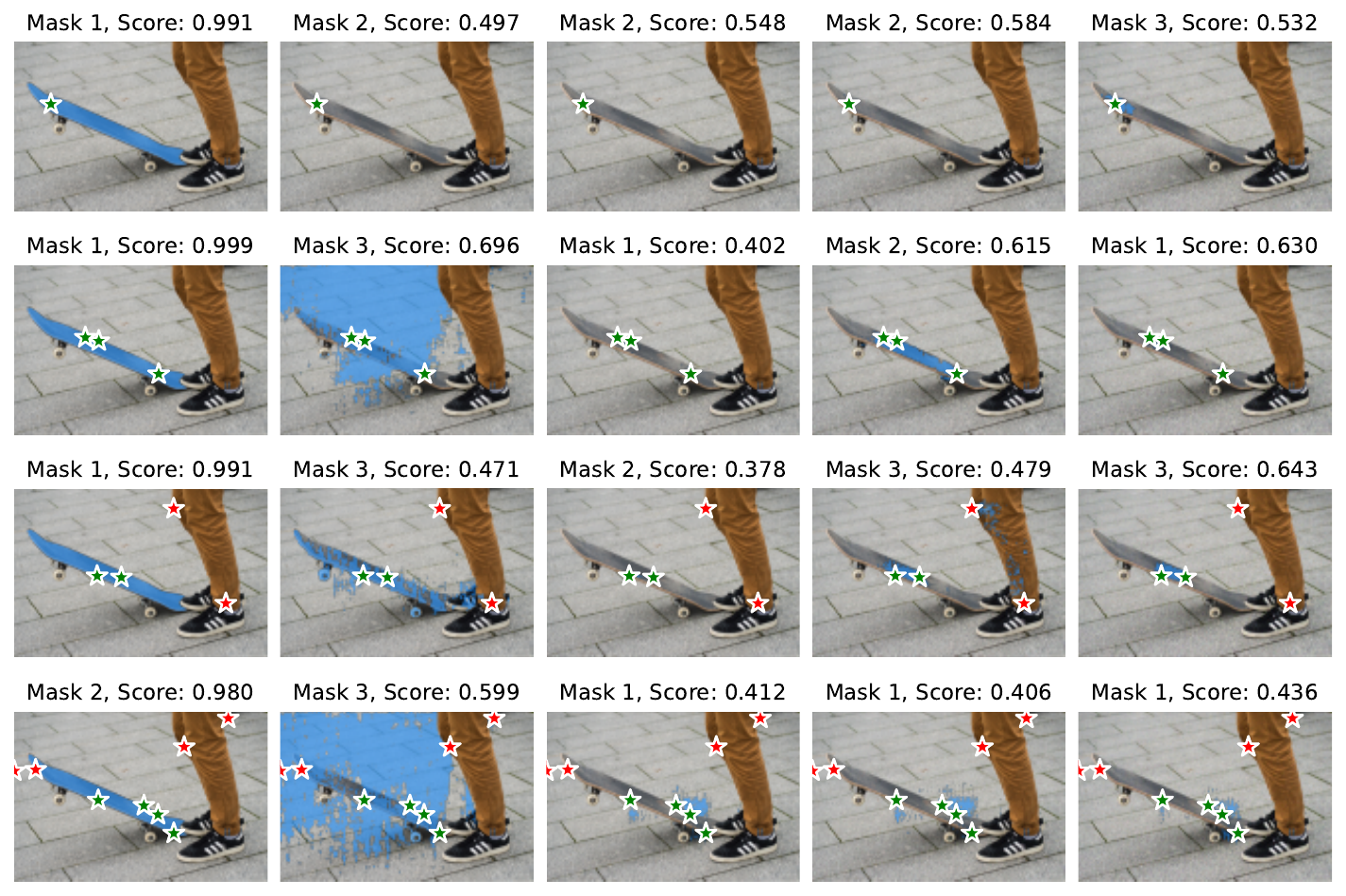}} & \multicolumn{5}{c}{\includegraphics[align=c, width=\newl]{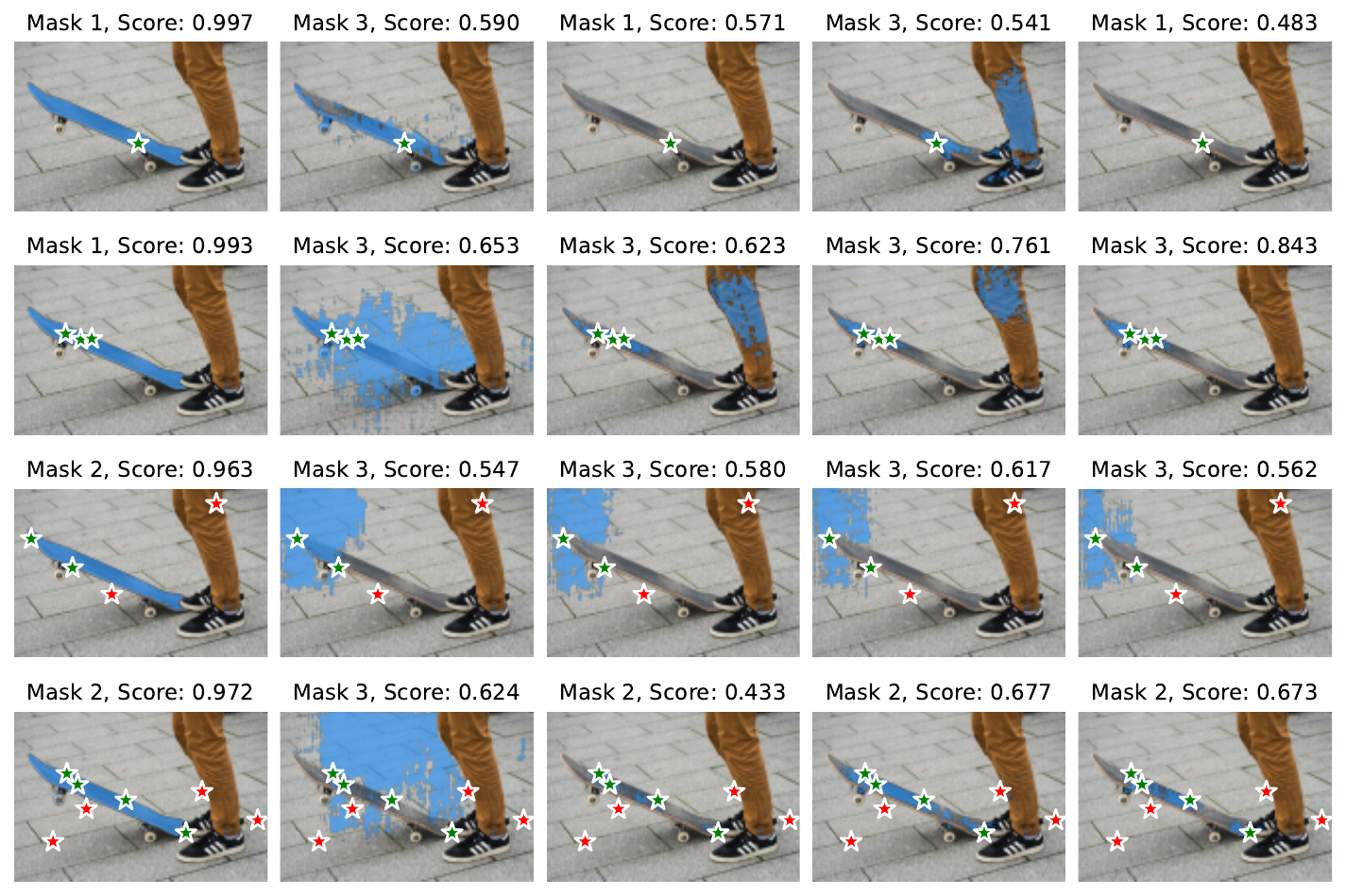}}

\end{tabular}
\caption{\textbf{Multiple point prompts on \sam.} We repeat the experiment in the setting reported in Fig.~\ref{fig:sam-multi-point-prompts} for a different input image from \saoneb, with similar observations about the effectiveness of the attacks.} \label{fig:sam_multi_point_prompts_2}

\end{figure*}

\textbf{Point prompts.}
Next we test the effect of the adversarial perturbations on \sam when using point prompts to segment a desired object. We explore using both positive and negative prompts (which indicate whether a point belongs or not to the target object).
Fig.~\ref{fig:sam-multi-point-prompts} and Fig.~\ref{fig:sam_multi_point_prompts_2} show the predicted masks with $p$ positive and $n$ negative prompts, for $(p, n) \in \{(1, 0), (3, 0), (2, 2), (4, 4)\}$, for several images, objects and seeds. In particular, we select masks from the annotations of \saoneb and uniformly sample points inside or outside it as positive and negative prompts respectively. Since \sam outputs three proposals as segmentation masks (see Fig.~\ref{fig:vis_sam}) with associated quality scores, we show the one with highest score.
We observe that, while for all prompts the original image is precisely segmented, the adversarially perturbed inputs, even small $\ell_\infty$-bounds lead to very different results, and in most cases the mask fails to identify the target object.
Increasing the number of prompts does not lead, in general, to recovering the correct segmentation, especially at the largest radius $\epsilon=8/255$.
We recall that the same perturbed image is used with all prompts and masks, which have not been seen by the attack when optimizing the perturbations.
\\

\begin{figure*} \centering \small
\newl=1.02\columnwidth
\tabcolsep=1.1pt 
\begin{tabular}{*{5}{C{17mm}} | *{5}{C{17mm}}}

original & $1/255$ & $2/255$ & $4/255$ & $8/255$ & original & $1/255$ & $2/255$ & $4/255$ & $8/255$ \\
\toprule

\multicolumn{5}{c|}{\includegraphics[align=c, width=\newl]{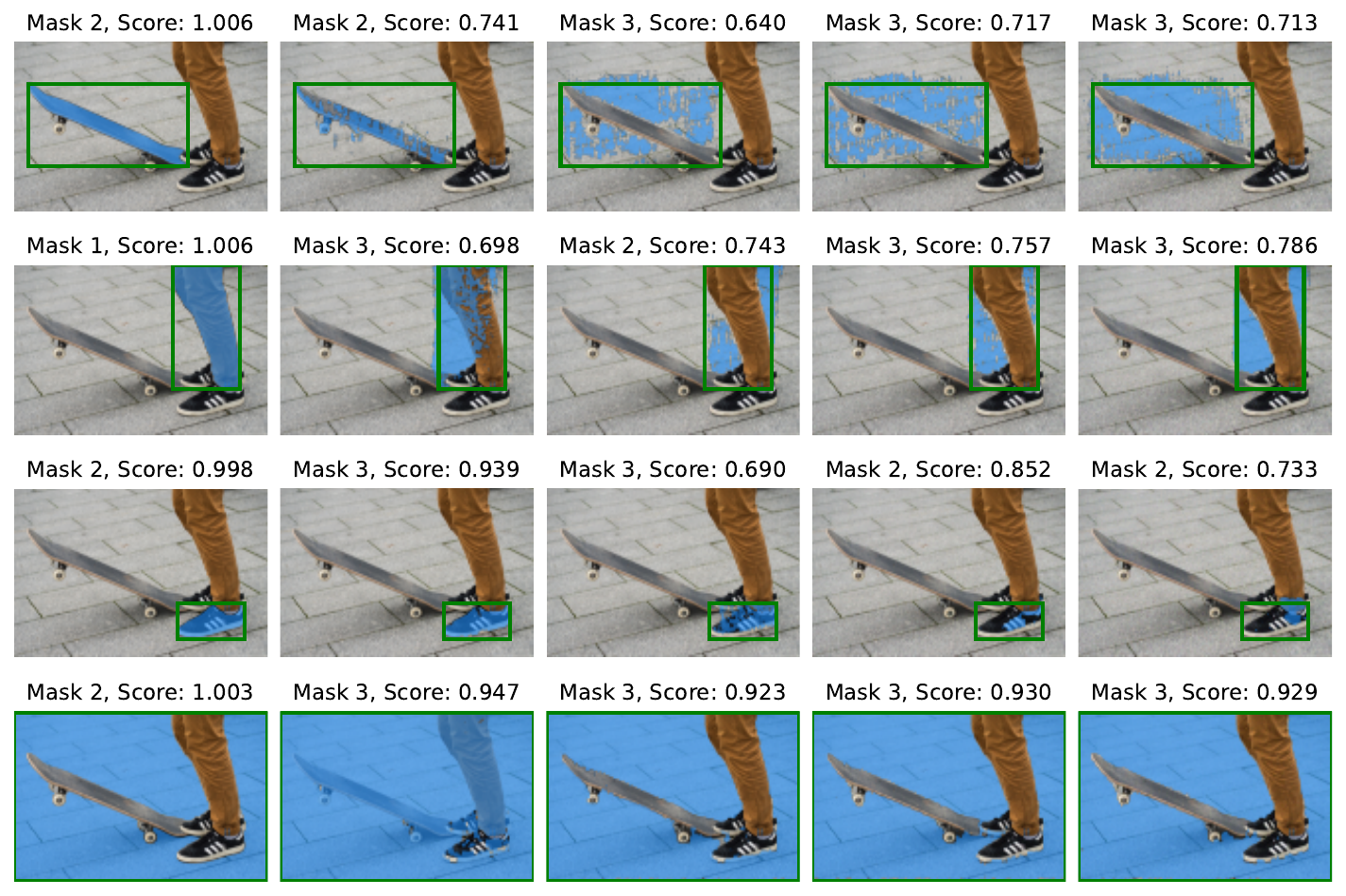}} & \multicolumn{5}{c}{\includegraphics[align=c, width=\newl]{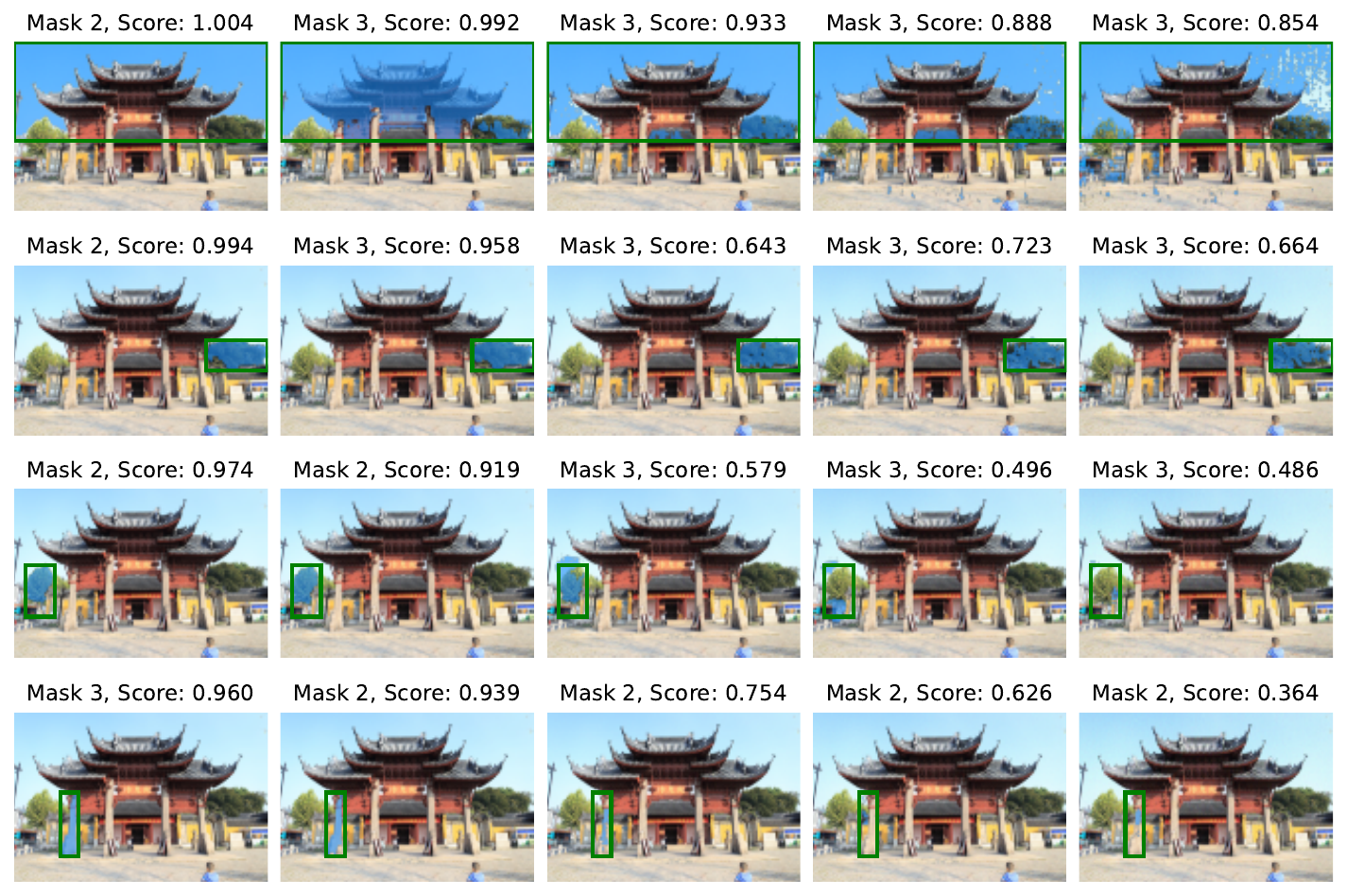}} \\
\midrule
\multicolumn{5}{c|}{\includegraphics[align=c, width=\newl]{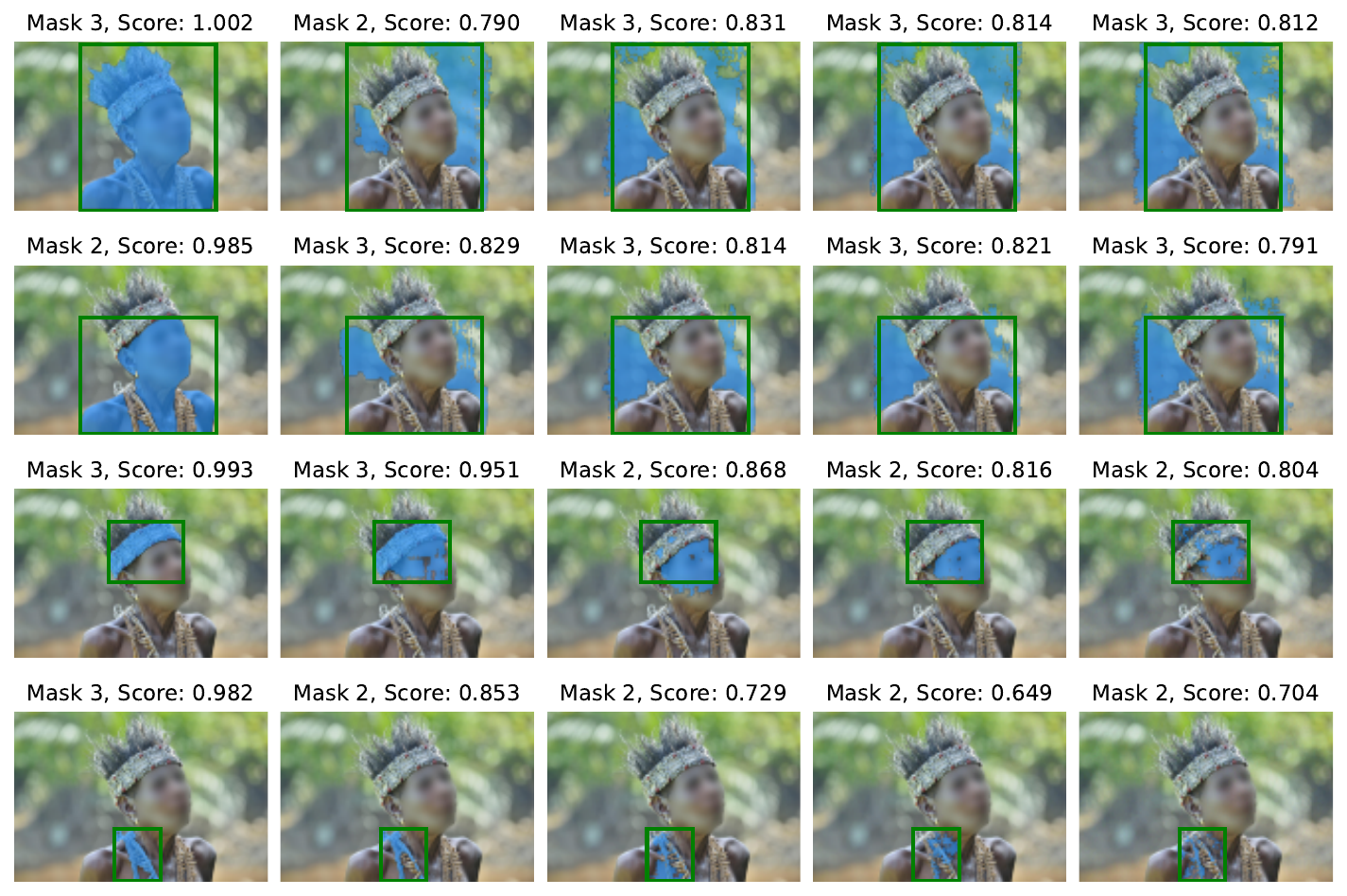}} & \multicolumn{5}{c}{\includegraphics[align=c, width=\newl]{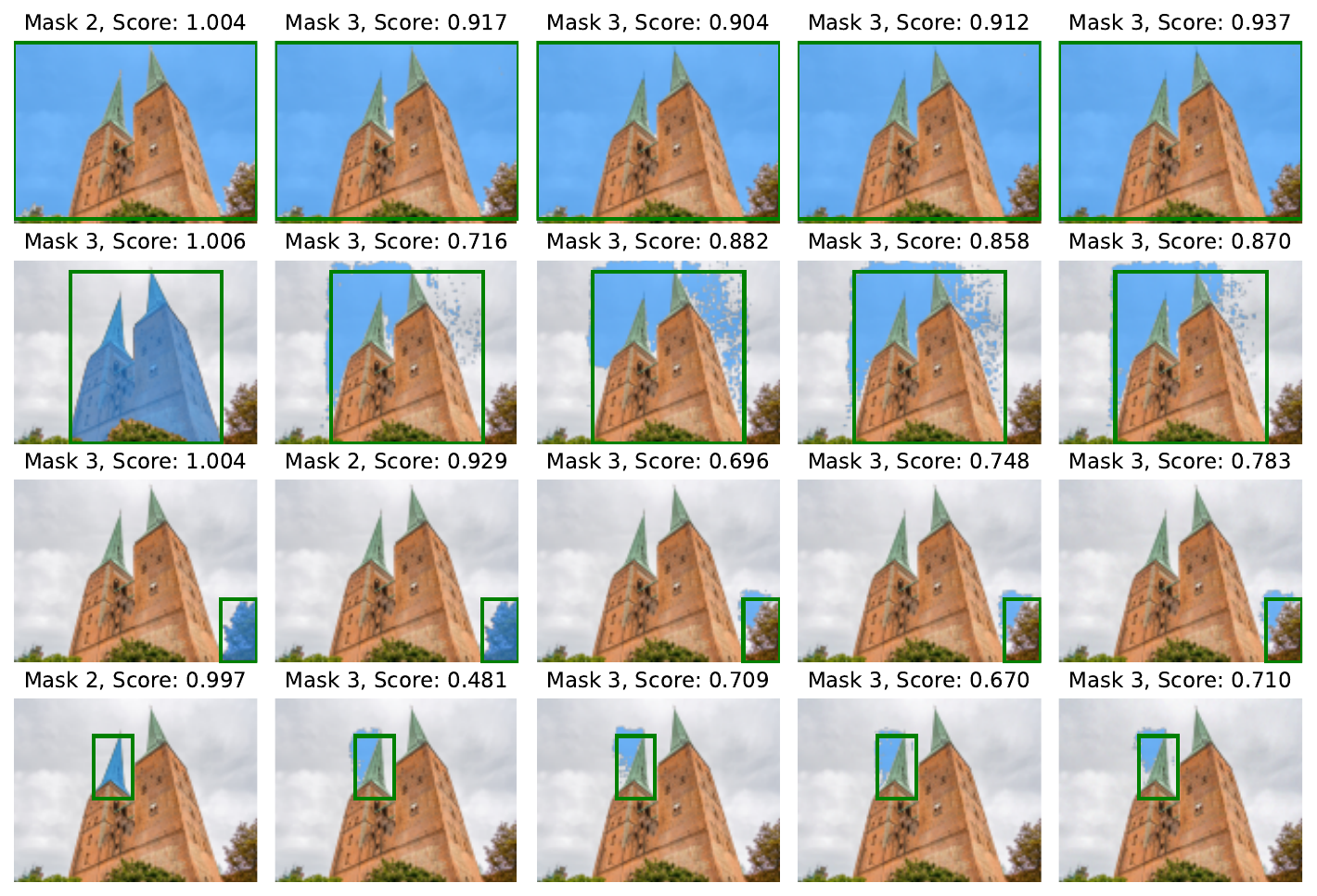}}

\end{tabular}
\caption{\textbf{Box prompts on \sam.} For each image we show the segmentation masks obtained with several box prompts and its quality score as in Fig.~\ref{fig:sam-multi-point-prompts}. We use either the original image or those perturbed with perturbations of $\ell_\infty$-norm $\epsilon\in\{1/255, 2/255, 4/255, 8/255\}$. Small perturbations effectively degrade the mask quality, especially for small and medium size objects.} \label{fig:sam_box_prompts}

\end{figure*}

\textbf{Box prompts.}
Similarly to point prompts, we select several box prompts from the dataset annotations and show the effect of using perturbed inputs (varying radii $\epsilon\in\{1/255, 2/255, 4/255, 8/255\}$) in combination with them in Fig.~\ref{fig:sam_box_prompts}.
The adversarial images often lead to masks which are either (almost) empty, inside the box but complementary to the correct ones or of poor quality, sometimes leaving the boundaries of the box.
Only when the box includes the entire image, which for unperturbed images results in the background being segmented, the attacks cannot completely erase the correct masks.

\subsection{Universal attacks on \sam} \label{sec:universal_attacks_sam}

\begin{figure*} \centering \small
\newl=.25\columnwidth
\colwidth=.5\columnwidth
\tabcolsep=1.1pt 
\begin{tabular}{*{1}{C{\colwidth}} | *{1}{C{\colwidth}}
|| *{1}{C{\colwidth}} | *{1}{C{\colwidth}}
}
\multicolumn{2}{c||}{\textbf{seen images}} & \multicolumn{2}{c}{\textbf{unseen images}} \\[2mm]

original & perturbed ($\epsilon=8/255$) & original & perturbed ($\epsilon=8/255$) \\
\toprule

\multicolumn{2}{c||}{\includegraphics[align=c, width=4\newl]{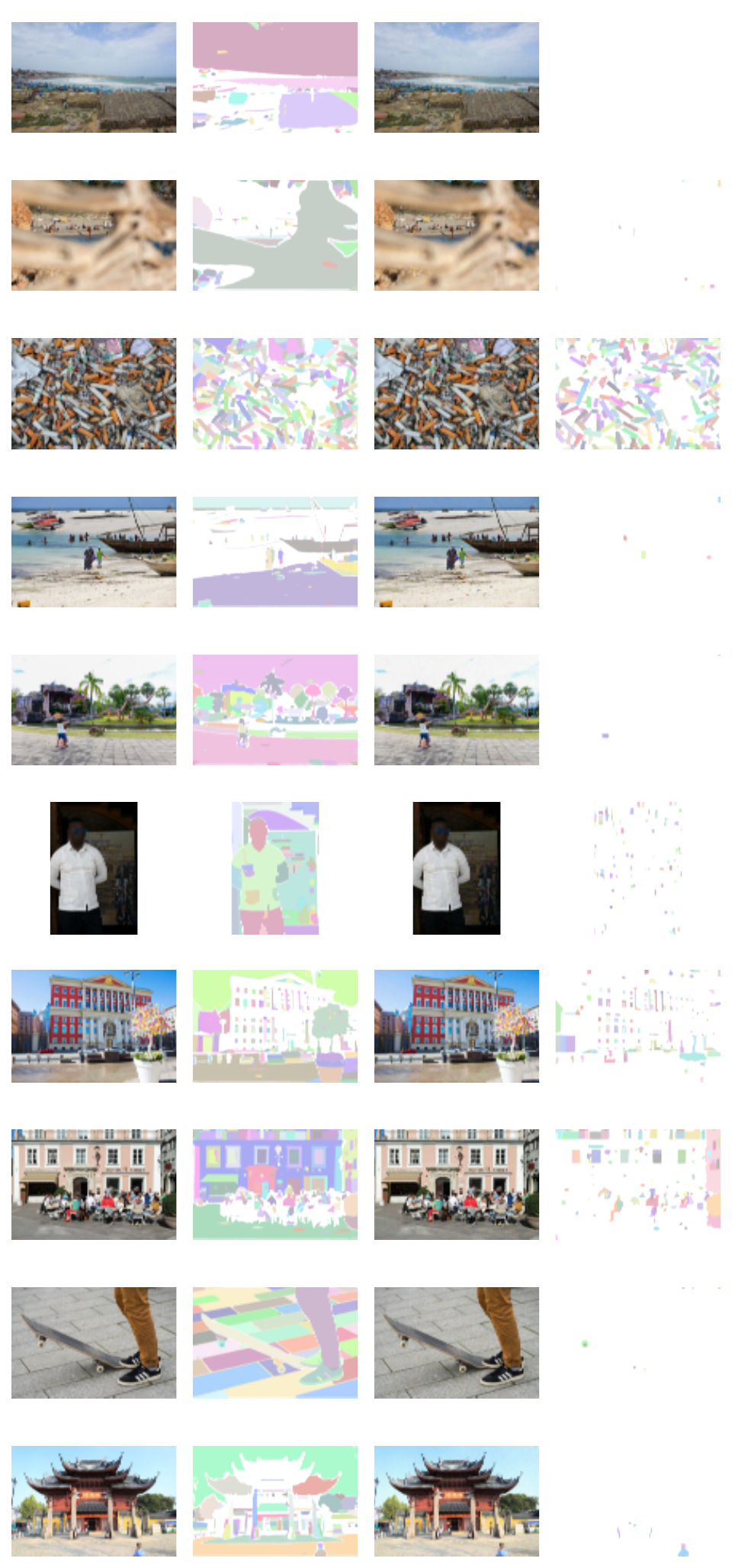}} & 
\multicolumn{2}{c}{\includegraphics[align=c, width=4\newl]{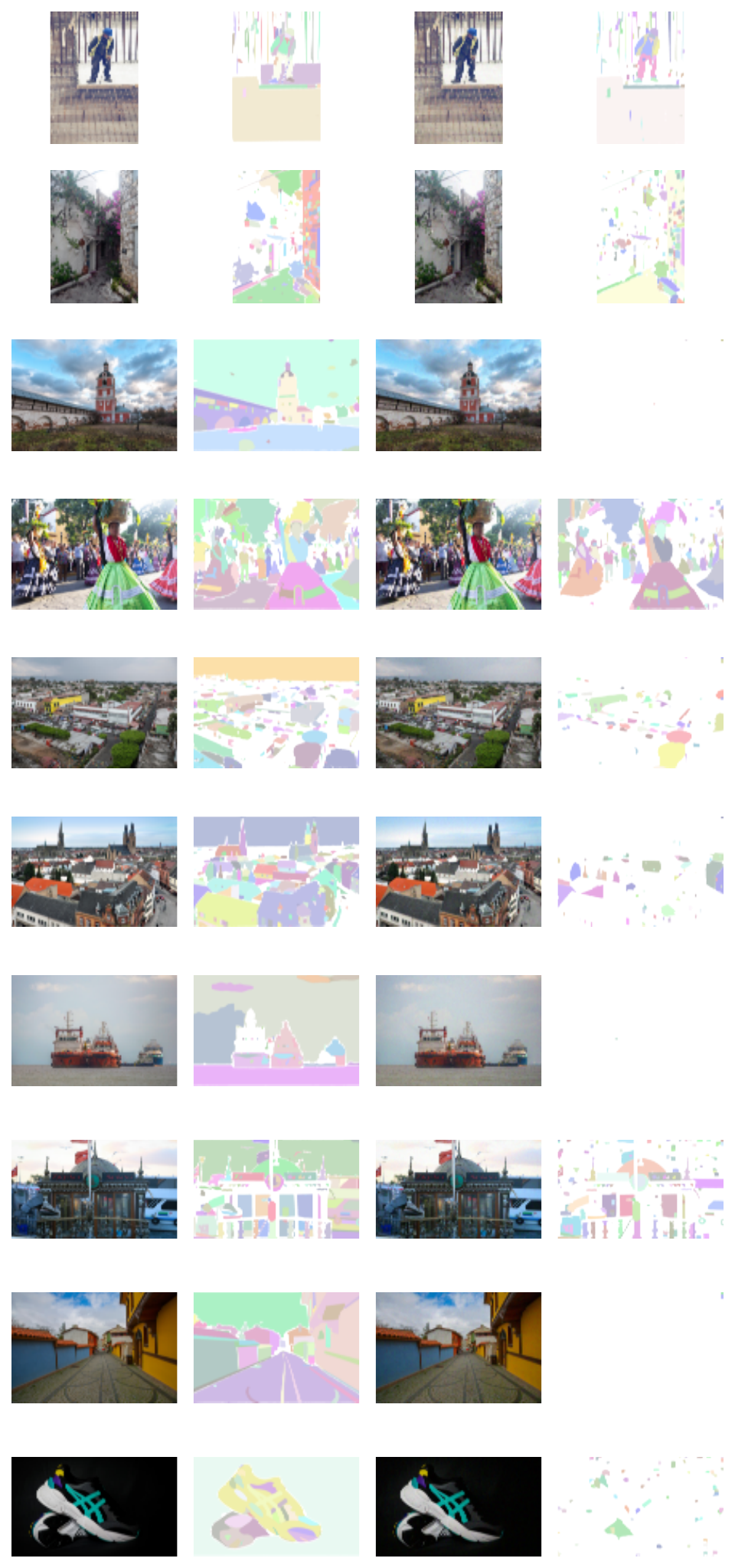}}

\end{tabular}
\caption{\textbf{Universal attacks on \sam.} In the case of universal attacks, the same perturbation (with $\ell_\infty$-norm of $8/255$) is applied to every image, and we study its effect on the Segment Everything mode of \sam.
On the left we show a subset of the images used for generating the universal attack (first column), together with their predicted segmentation masks (second), the images obtained after adding the universal perturbation (third) and their corresponding predictions (fourth).
On the right we follow the same procedure, this time using images not seen by the attack. While some areas are still correctly segmented, for most images the adversarial attacks either prevents any precise segmentation or introduce many small masks.}\label{fig:sam_univ_attacks}
\end{figure*}

While we have so far tested image-specific perturbations, we now aim at finding a \textit{single} perturbations which can be applied to any input image and prevent generating precise segmentation masks for it.
This corresponds to modifying Eq.~\eqref{eq:prompt-agnostic-attacks} to
\begin{align} \begin{split} \maxop_{\vdelta \in \R^{w\times h \times c}} \; &\sum_{i=1}^n 
\, \norm{\phi(\vx_i + \vdelta)-\phi(\vx_i)}_2^2
\\
&\textrm{s.th.} \quad \norm{\vdelta}_p \leq \epsilon, \quad \vx_i + \vdelta \in [0, 1]^{w\times h \times c},
\end{split} \label{eq:prompt-agnostic-univ-attacks}
\end{align}
where one jointly optimizes the loss for $n$ training images, and the same perturbation $\vdelta$ is added to all of them.

However, the training images, and more importantly the test images, will not in general have the same resolution. Thus we fix the shape of the perturbation
to 1024x1024 pixels (note that the smaller side of each image in \saoneb is 1500 pixels), and interpolate it to match the size of the target image via a function\footnote{In practice we use the function available in \texttt{torch} with default ``nearest'' mode, see  details at \url{https://pytorch.org/docs/stable/generated/torch.nn.functional.interpolate.html}.} $g$. This changes the optimization problem in Eq.~\eqref{eq:prompt-agnostic-univ-attacks} to
\begin{align} \begin{split} \maxop_{\vdelta \in \R^{w\times h \times c}} \; &\sum_{i=1}^n 
\norm{\phi\big(\vx_i + g(\vdelta)\big)-\phi(\vx_i)}_2^2
\\
&\textrm{s.th.} \quad \norm{\vdelta}_p \leq \epsilon, \quad \vx_i + g(\vdelta) \in [0, 1]^{w_i\times h_i \times c},
\end{split} \label{eq:prompt-agnostic-univ-attacks-interpolation}
\end{align}
where the image $\vx_i$ has resolution $w_i\times h_i$.
When optimizing the attack, we compute the gradient of the target loss wrt each input image, normalize it wrt its $\ell_2$-norm (so that all images have comparable influence on the updates), and finally sum them.

We select a random set of 100 images for generating the attack. To avoid overfitting to such training images, at each iteration we randomly sample a batch of 10 out of the 100 training images and make a gradient step to optimize the sum of their losses. This procedure is also meant to foster generalization to unseen images. We use PGD as optimizer with $\epsilon=8/255$, step size $1/255$ and 500 iterations, which amounts to 5000 total gradient computations (since this is a more challenging setup we use larger perturbation size and computational budget).

In Fig.~\ref{fig:sam_univ_attacks} we show the effect of the found universal attacks on the results of the Segment Everything mode of \sam (see Sec.~\ref{sec:point_prompts_ade} for an evaluation of universal attacks with single point prompts). For both training (seen) and unseen images, adding the perturbation significantly deteriorates the predicted segmentation masks. 
For the majority of cases, when the adversarial attack is applied, almost no mask is produced, and sometimes many small segmentation masks (not corresponding to any object in the image) appear.
We further show the results in the same setup with either a different set of randomly selected training images (Fig.~\ref{fig:sam_univ_attacks_seed-0}) or the smaller perturbation budget $\epsilon=4/255$ (Fig.~\ref{fig:sam_univ_attacks_eps_4}) in App.~\ref{sec:additional_results_app}.
Although these universal attacks produce slightly worse degradation than the image-specific ones (see Fig.~\ref{fig:sam_segment_everything}, where even smaller radii up to $4/255$ are used), they are realized with a single perturbation which can be applied to any input image.
Finally, our goal in this context was to show that prompt-agnostic universal attacks can be achieved even from a small set of images, and we expect that allocating more computational budget for the algorithms, e.g. using more training images, iterations or larger batch size, can improve the generalization of universal attacks.

\subsection{Evaluation on \ade masks with single point prompts} \label{sec:point_prompts_ade}

\begin{table} \centering
\small \newl=10mm
\tabcolsep=2.5pt
\extrarowheight=1.5pt
\caption{\textbf{Single point prompt evaluation on \sam.} We study the performance of \sam when predicting masks derived from \ade with single point prompts (we measure average IoU over images and masks). We report the effect of using either image-specific or universal (trained on images of \ade not used for computing mIoU) attacks of various sizes. 
Both types of adversarially perturbed images lead to significant performance drops.
}
\label{tab:evaluation_ade}
\begin{tabular}{C{20mm} 
|| C{\newl} | *{4}{C{\newl}}}
attack & clean & $1/255$ & $2/255$& $4/255$ & $8/255$ \\
\toprule
image-specific 
& \multirow{2}{*}{59.86} & 16.71 & 11.63 & 9.42 & 7.33 \\
universal 
&  & 58.98 & 45.09 & 20.08 & 11.24 \\
\bottomrule
\end{tabular}
\end{table}

In this section, we provide a quantitative evaluation of the performance degradation of \sam due to our attacks.\\

\textbf{Experimental setup.}
First, we create a set of ground truth masks and corresponding prompts from the \ade dataset~\cite{zhou2019semantic}, since it contains precise annotations and its images are typically evaluated at relatively low resolution (512x512), which allows us to scale the evaluation to a larger number of images and masks.
Given an image from the validation set, for each class present in the ground truth semantic segmentation map (except for the background) we select the largest connected component belonging to such class. For each of these masks, after filtering out those with area smaller than 900 pixels, we compute its pixel with largest $\ell_2$-distance to its border, following \cite{kirillov2023segany}.
In this way we collect 100 images and 589 pairs of masks and point prompts.

Then, we can use \sam to predict segmentation masks from each of the point prompts found in the step described above and either the original image or its adversarially perturbed counterparts. 
Comparing the predicted masks with the ground truth ones derived from \ade allows us to measure the performance of both \sam on the unperturbed image and the effectiveness of the attacks.
In practice, for each mask-prompt pair we compute Intersection over Union of the predicted (the one with highest quality score among the three provided by \sam) and correct masks, and report its average over all masks and images (mIoU) in Table~\ref{tab:evaluation_ade}.
\\

\textbf{Results.} First, we run our prompt-agnostic image-specific attack with radii $\epsilon\in\{1/255, 2/255, 4/255, 8/255\}$, optimized with 100 iterations APGD, on the 100 images used to create the evaluation set.
Second, we generate universal perturbations as described in Sec.~\ref{sec:universal_attacks_sam}, with 100 iterations of PGD (instead of the 500 iterations used in the previous section to reduce computational cost), for the same radii as image-specific attacks. 
In this case, as training images, we select 100 samples from \ade which do not overlap with those used for evaluation (which means the universal attacks are tested on unseen images), and the universal perturbation has resolution 512x512 as the images of \ade.

In Table~\ref{tab:evaluation_ade} we see that, for image-specific attacks, even perturbations of size $1/255$ are able to significantly reduce mIoU from 59.86\% to 16.71\%. Increasing the attack budget to $8/255$ further degrades the performance of \sam to 7.33\%.
Conversely, for universal attacks one needs to use the larger radii for effective attacks: notably, at $\epsilon = 8/255$, the mIou for adversarial images is 11.24\%, not far from what attained with standard attacks (we recall that the universal perturbations are computed only once and then applied to unseen images without additional cost).

\subsection{Adaptation to different mask generators in \sam} \label{sec:multicrop_generator}

\begin{figure*} \centering \small
\newl=.24\columnwidth \rowheight=75mm
\tabcolsep=1.3pt 
\begin{tabular}{L{4mm} *{2}{C{\newl}}| *{3}{C{\newl}} | *{3}{C{\newl}}}

 & & & \multicolumn{3}{c|}{\textbf{standard APGD}} & \multicolumn{3}{c}{\textbf{multi-crop PGD}} \\[2mm]

 & & original & $1/255$ & $2/255$& $4/255$ & $1/255$ & $2/255$ & $4/255$  \\
\toprule

\rotatebox[origin=c]{90}{\textbf{standard generator}} &\multicolumn{2}{c|}{\includegraphics[align=c, height=\rowheight]{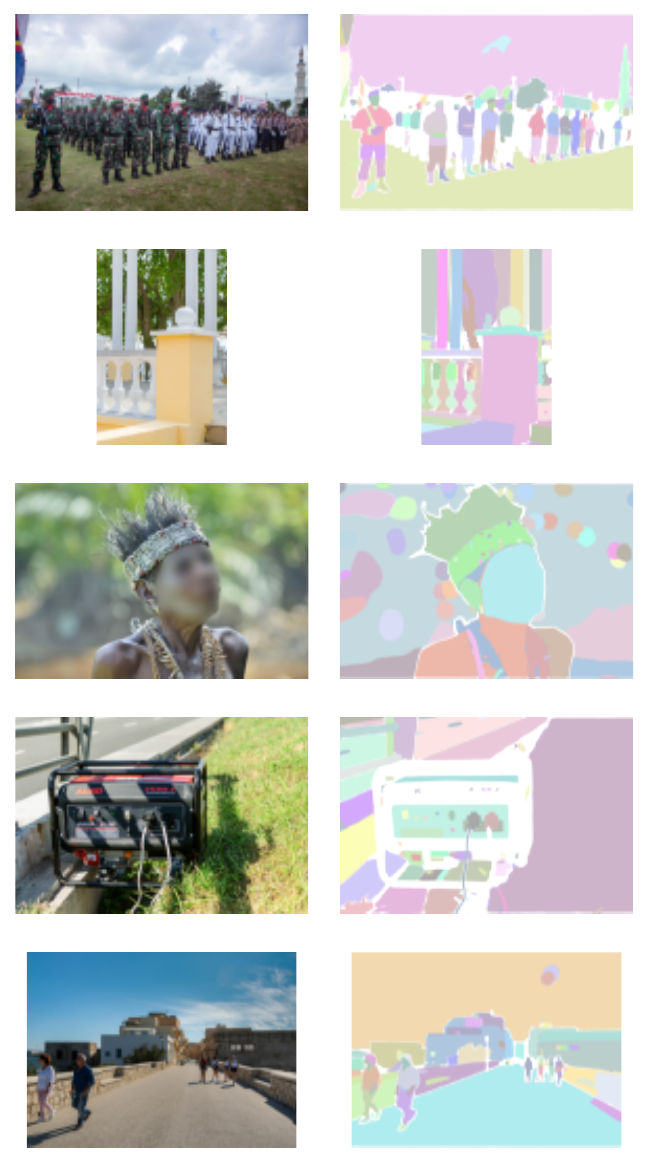}} &
\multicolumn{3}{c|}{\includegraphics[align=c, height=\rowheight]{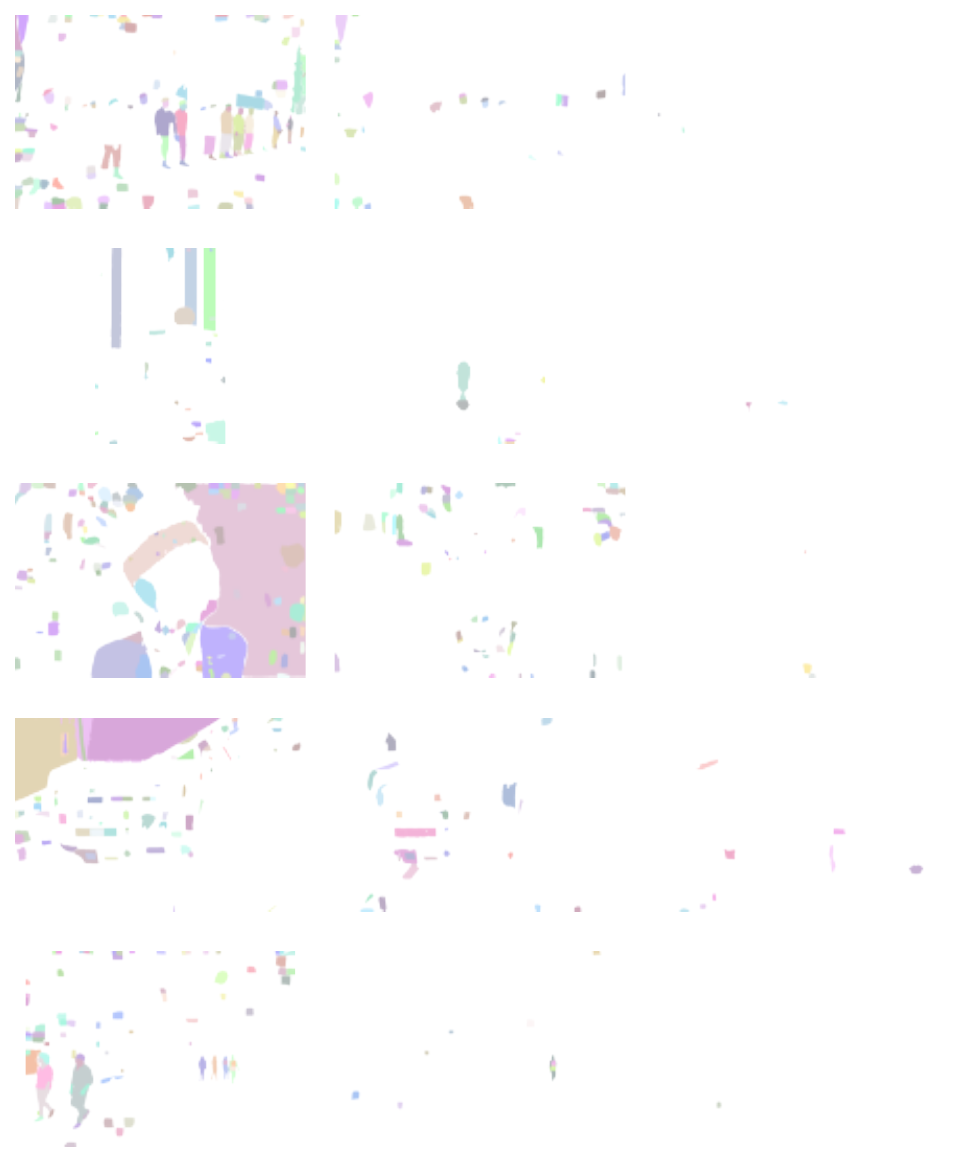}} &
\multicolumn{3}{c}{\includegraphics[align=c, height=\rowheight]{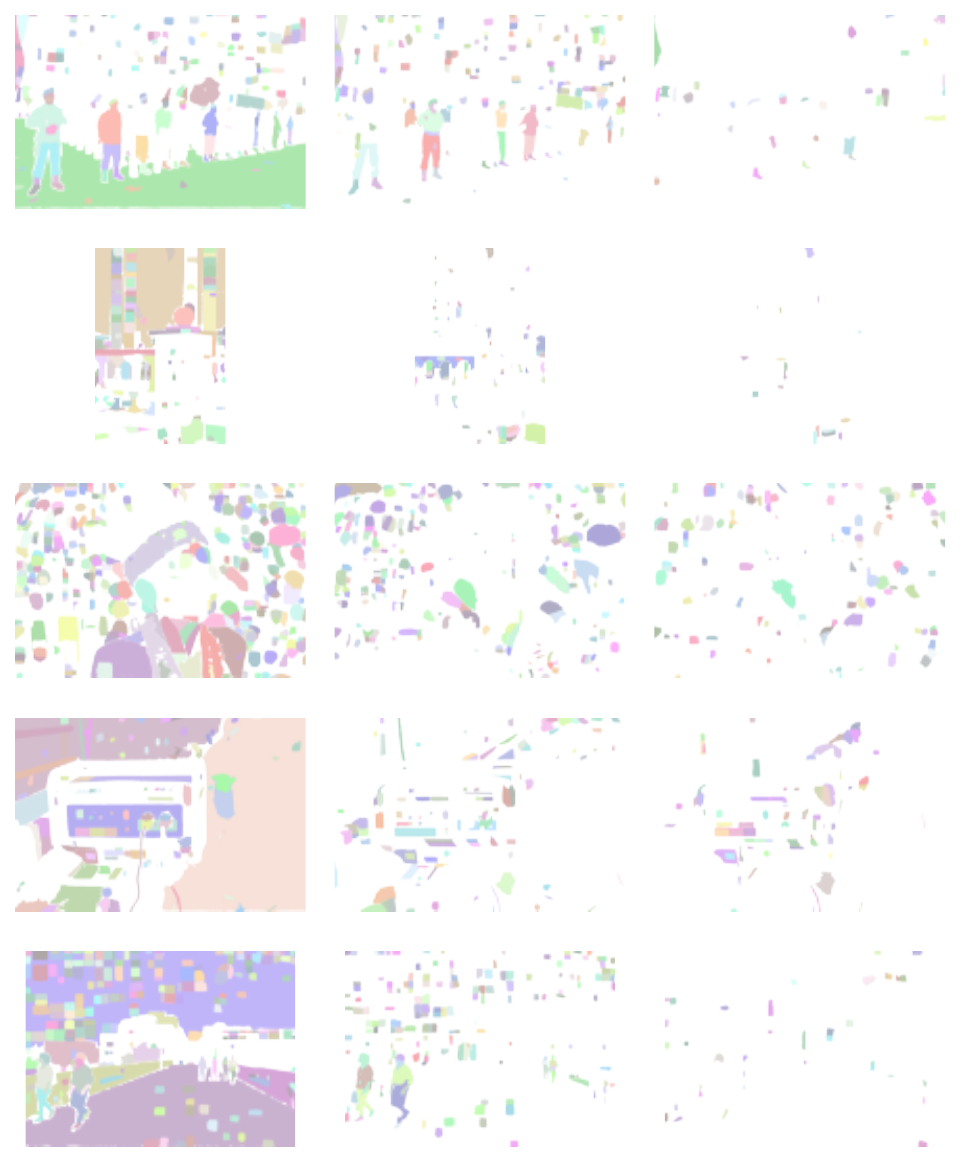}}\\
\midrule

\rotatebox[origin=c]{90}{\textbf{multi-crop generator}}&\multicolumn{2}{c|}{\includegraphics[align=c, height=\rowheight]{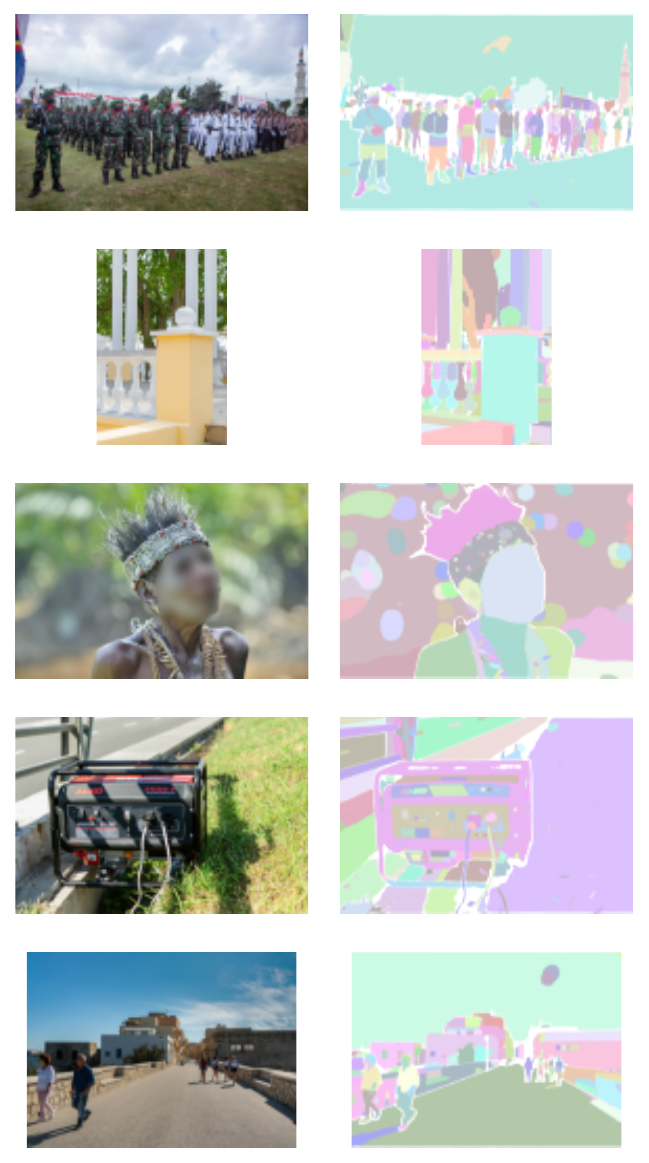}} &
\multicolumn{3}{c|}{\includegraphics[align=c, height=\rowheight]{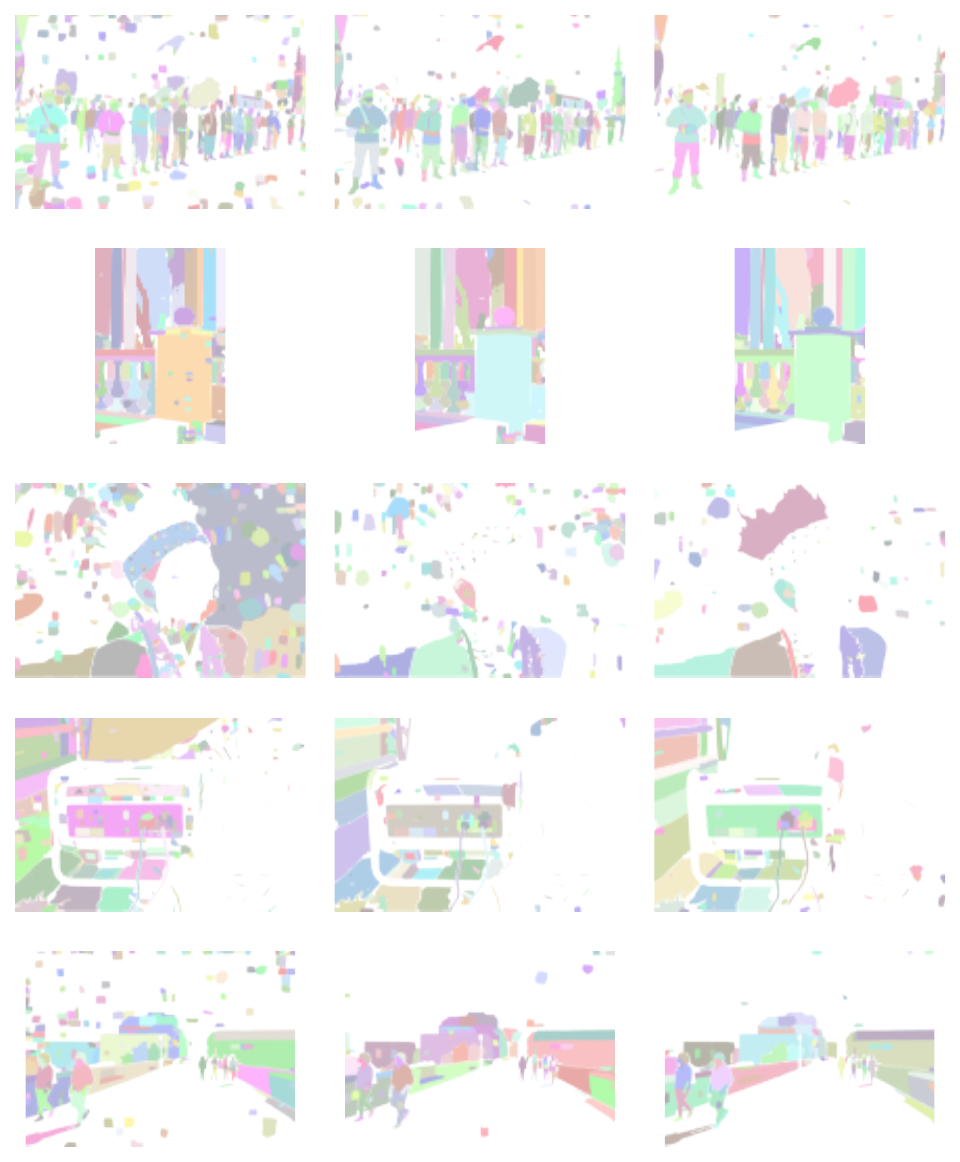}} &
\multicolumn{3}{c}{\includegraphics[align=c, height=\rowheight]{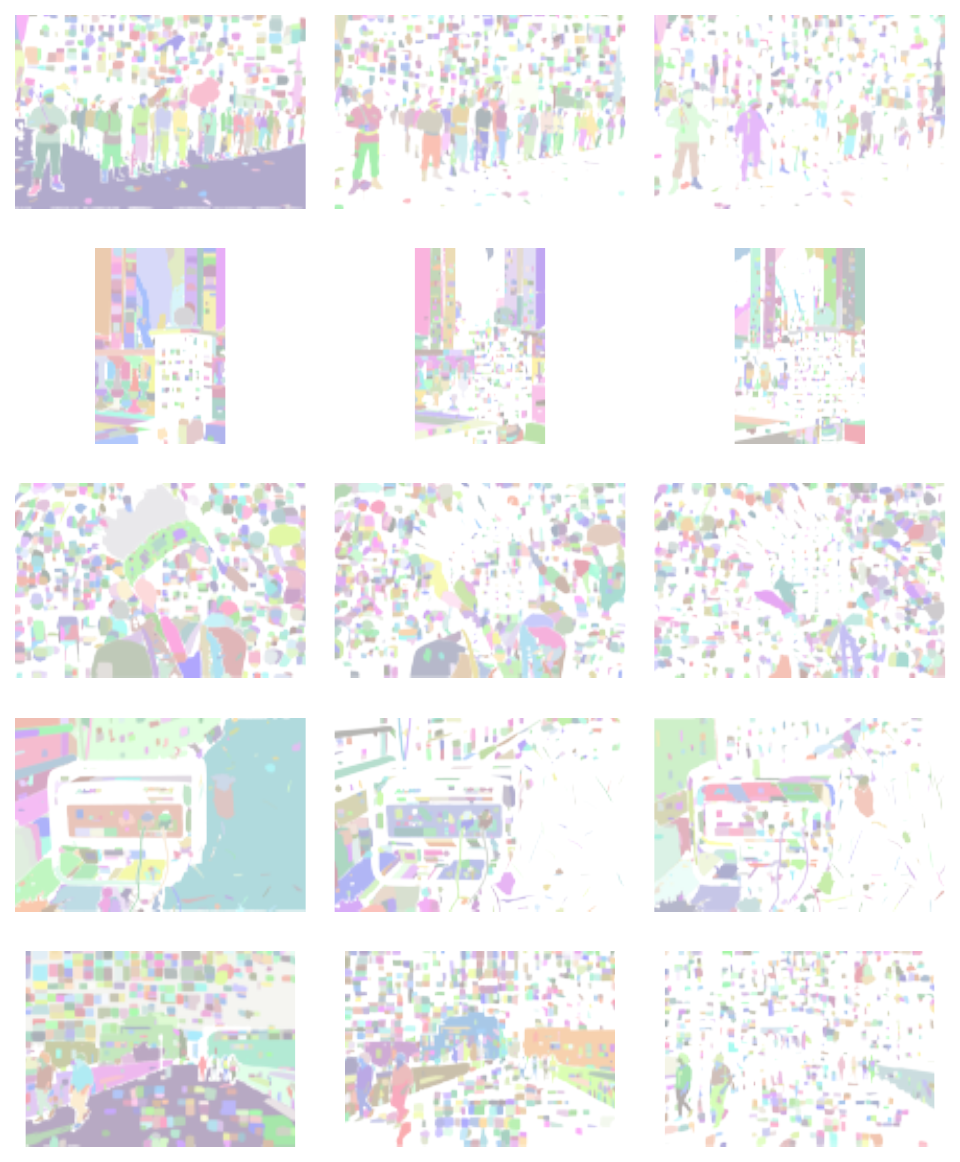}}

\end{tabular}
\caption{\textbf{Effect of different mask generators and attack algorithms on \sam.} We show the mask predicted by \sam (Segment Everything mode) with standard (top part) and multi-crop (bottom) mask generators, for the clean images (left column) and the adversarially perturbed image produced by either standard APGD (middle) and multi-crop PGD (right). We use each attack with three perturbation bounds $\epsilon \in \{1/255,2/255, 4/255\}$.
The multi-crop PGD is also effective when the multi-crop generator is used, with better results than APGD.
} \label{fig:sam_multicrop_vs_standard_generator}
\end{figure*}

As an option to control some properties of the predicted masks in the Segment Everything mode, it is possible to tune several hyperparameters of the mask generator in \sam.
In particular, a more sophisticated mask generator, which computes the segmentation masks for multiple crops of the image and then combines them, is suggested in the original code.\footnote{The details of the parameters used by the mask generators are reported in \url{https://github.com/facebookresearch/segment-anything/blob/main/notebooks/automatic_mask_generator_example.ipynb}.}
While this multi-crop generator is more computationally expensive than the default one, it might produce more refined results, as shown in Fig.~\ref{fig:sam_multicrop_vs_standard_generator}.
Note that this is independent of the image encoder, so different configurations of the generator do not affect the optimization of prompt-agnostic attacks, and can be used to test them.

In Fig.~\ref{fig:sam_multicrop_vs_standard_generator} we first show the performance of both standard (the one also used in the previous experiments) and multi-crop generators on clean images (left column) and on the adversarially perturbed inputs given by APGD (middle column) with radii $\epsilon \in \{1/255, 2/255, 4/255\}$ (the same attacks used for Fig.~\ref{fig:sam_segment_everything}). We see that while the attacks are very effective with the standard generator, the multi-crop configuration is, in some cases, still able to segment several objects, especially small ones.
We hypothesize that this is due to the fact that in the standard APGD all pixels of the perturbation contribute to the distortion of the features given by the image encoder. However, when only a smaller portion of the image is used, the cropped perturbation might not be as effective.

To counter this, we design a new algorithm, named multi-crop PGD, where, at each iteration, with probability $p_\textrm{crop}=0.8$ we use a randomly cropped version of the image to compute the objective loss and update the current perturbation.
In particular, we first select a random rectangular subset of the current iterate (the perturbed image at the current iteration) whose width and height are uniformly (and independently one from another) sampled between $30\%$ and $90\%$ of the original width and height respectively. Then we make an update step to maximize the feature distortion for this cropped image.
Note that this updates the adversarial perturbation only in the area corresponding to the sampled crop.
We use 100 iterations of PGD with step size $\epsilon / 8$ (we do not use APGD since the objective function is not the same for all iterations).

In the right part of Fig.~\ref{fig:sam_multicrop_vs_standard_generator} we show the results of multi-crop PGD: when using the standard mask generator its attacks are still effective, although slightly less than those of standard APGD. At the same time, it is able to deteriorate the predicted masks of the multi-crop generator more significantly than APGD.
In particular, with both mask generators, multi-crop PGD leads to a large number of very small masks, with some similarities to what happens for the universal attacks (Sec.~\ref{sec:universal_attacks_sam}) and unlike the standard attacks.

\subsection{White-box attacks on \seem} \label{sec:white-box_seem}

Another recently proposed promptable segmentation model is \seem \cite{zou2023segment}, whose structure at high level resembles that of \sam with an image encoder that extracts, from an input image, features which can be then combined with various types of visual, text and audio prompts to solve various tasks.
In particular, \cite{zou2023segment} uses different backbones for the image encoder, and in the following we consider the Focal-T and Focal-L \cite{yang2022focal} which are publicly available.\footnote{We use the implementation and models provided at \url{https://github.com/UX-Decoder/Segment-Everything-Everywhere-All-At-Once/tree/main}.}
As above, we solve Eq.~\eqref{eq:prompt-agnostic-attacks} with 100 steps of APGD \cite{croce2020reliable} wrt $\ell_\infty$. As for \sam, the attacks only aim at perturbing the features generated by the image encoder (only the visual backbone is used) and do not consider the segmentation head or prompt encoders.
We use images from \ade and \saoneb: when using the larger backbone Focal-L, we resize the high resolution images from \saoneb to be able to run the attacks with batch size equal 1 in memory of a single GPU. In particular, we resize the images to have smallest edge of size 512 pixels, as suggested in the original code.\footnote{See \url{https://github.com/UX-Decoder/Segment-Everything-Everywhere-All-At-Once/blob/main/demo_code/tasks/interactive.py}. We use bilinear interpolation to preserve the elements of the images in [0, 1].}
\\

\begin{figure*} \centering \small
\newl=1.8\columnwidth
\tabcolsep=1.1pt 
\begin{tabular}{*{5}{C{.155\columnwidth}} | *{5}{C{.205\columnwidth}}}

\multicolumn{5}{c|}{\textbf{\ade}} & \multicolumn{5}{c}{\textbf{\saoneb}} \\[2mm]

& original & $1/255$ & $2/255$ & $4/255$ 
& & original & $1/255$ & $2/255$ & $4/255$ \\
\toprule

\multicolumn{5}{c|}{\includegraphics[align=c, height=\newl]{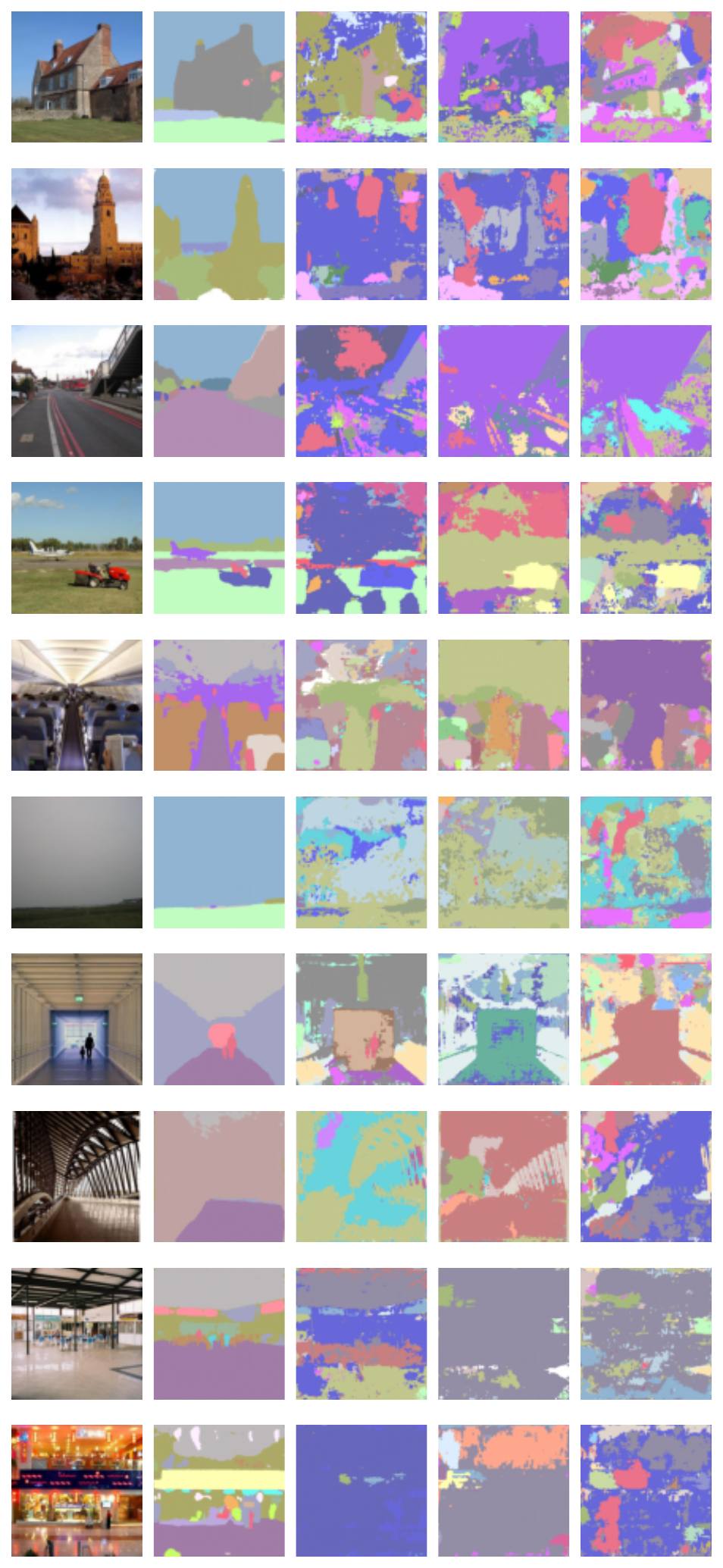}}&
\multicolumn{5}{c}{\includegraphics[align=c, height=\newl]{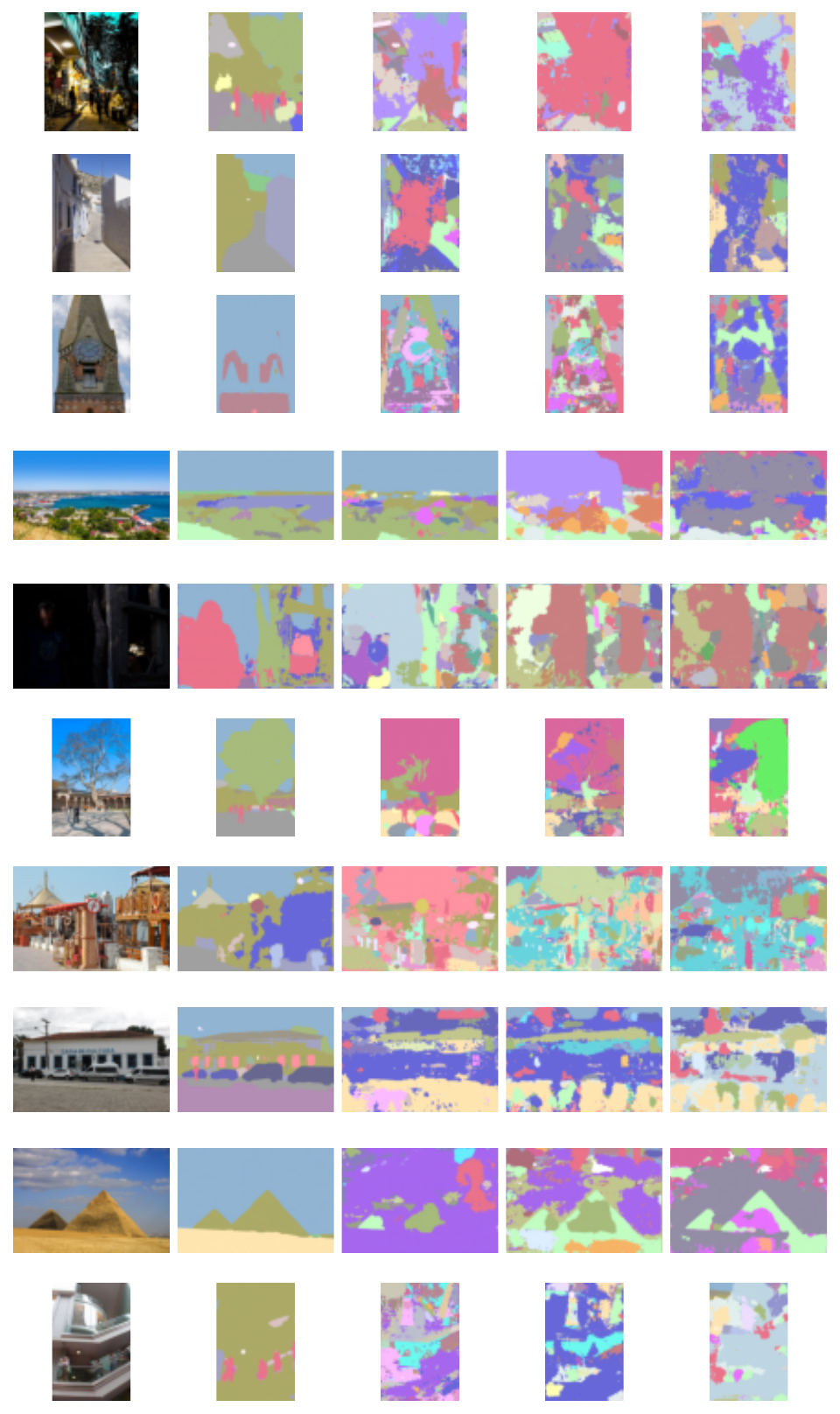}}

\end{tabular}
\caption{\textbf{Semantic segmentation with \seem (Focal-L).} For random images from the \ade (left) and \saoneb (right) dataset, we show the predicted semantic segmentation maps (each color corresponds to a predicted class) for the original input and the adversarially perturbed, with increasing radii, ones. Small perturbations are sufficient to drastically change the semantic segmentation maps predicted by \seem.} \label{fig:seem_focall_semantic}
\end{figure*}

\textbf{Semantic segmentation.}
Unlike \sam, \seem provides semantic segmentation maps for the input image, relying by default on the 133 classes (plus a void class) from \coco panoptic segmentation dataset \cite{lin2014microsoft}.
In Fig.~\ref{fig:seem_focall_semantic} we show the semantic segmentation masks predicted by \seem with Focal-L backbone for images from \ade and \saoneb and the corresponding adversarially perturbed versions given by APGD with $\ell_\infty$-bounds of $\epsilon\in \{1/255, 2/255, 4/255\}$.
We see that the even the small perturbations of size $1/255$ are sufficient to change the predicted classes. While some of the original shapes are still recognizable at the smallest thresholds, these disappear when increasing the budget of the attacks.
A similar experiment for the Focal-T backbone is shown in Fig.~\ref{fig:seem_focalt_semantic} in App.~\ref{sec:additional_results_app}, where the attacks achieve similar results to those on the larger image encoder. 
\\

\begin{figure*} \centering \small
\newl=1.02\columnwidth
\tabcolsep=1.1pt 
\begin{tabular}{@{} *{5}{C{17mm}} |  *{5}{C{17mm}}}

prompt & original & $1/255$ & $2/255$ & $4/255$ &  prompt & original & $1/255$ & $2/255$ & $4/255$  \\
\toprule
\multicolumn{5}{c|}{\includegraphics[align=c, width=\newl]{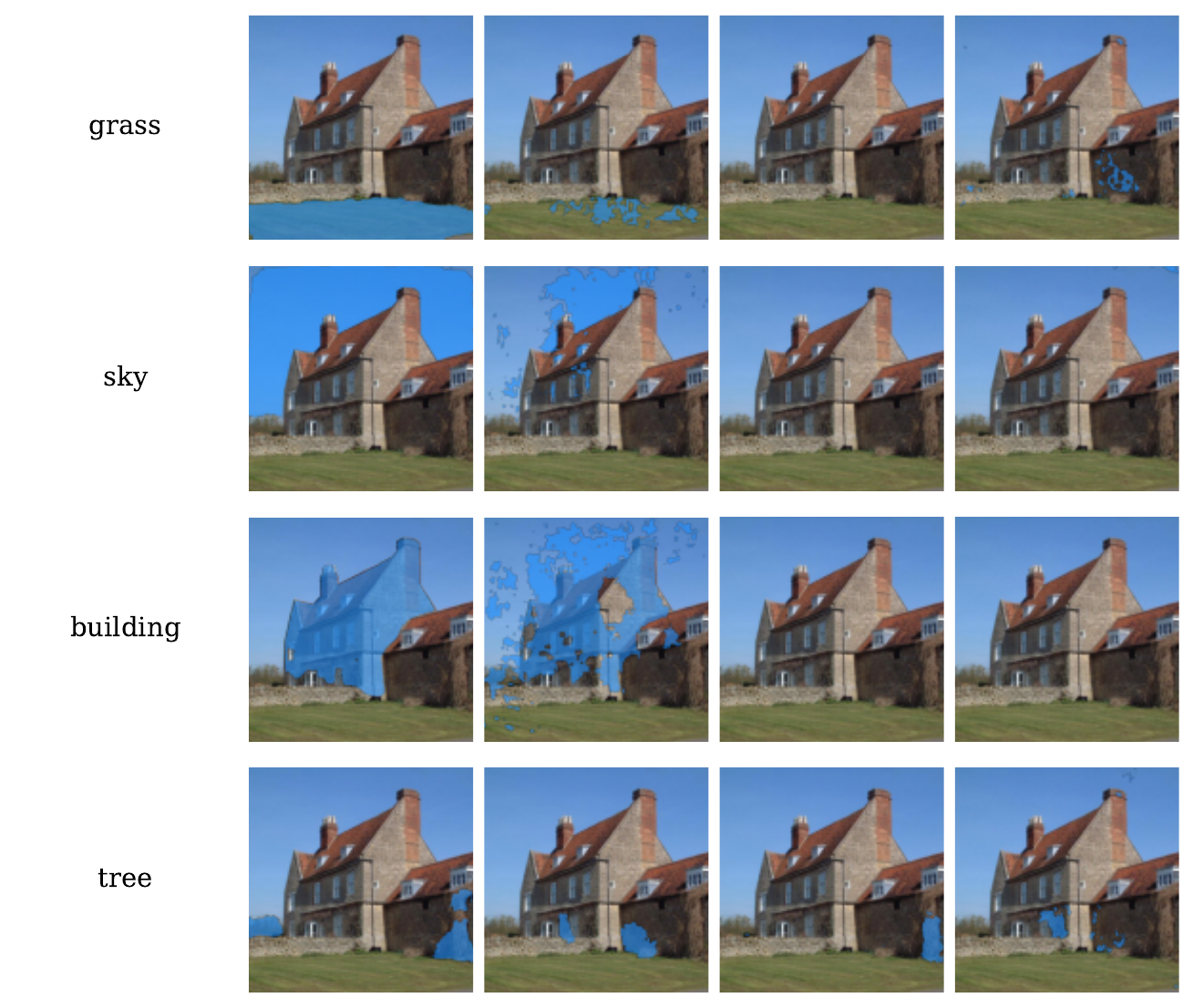}} & 
\multicolumn{5}{c}{\includegraphics[align=c, width=\newl]{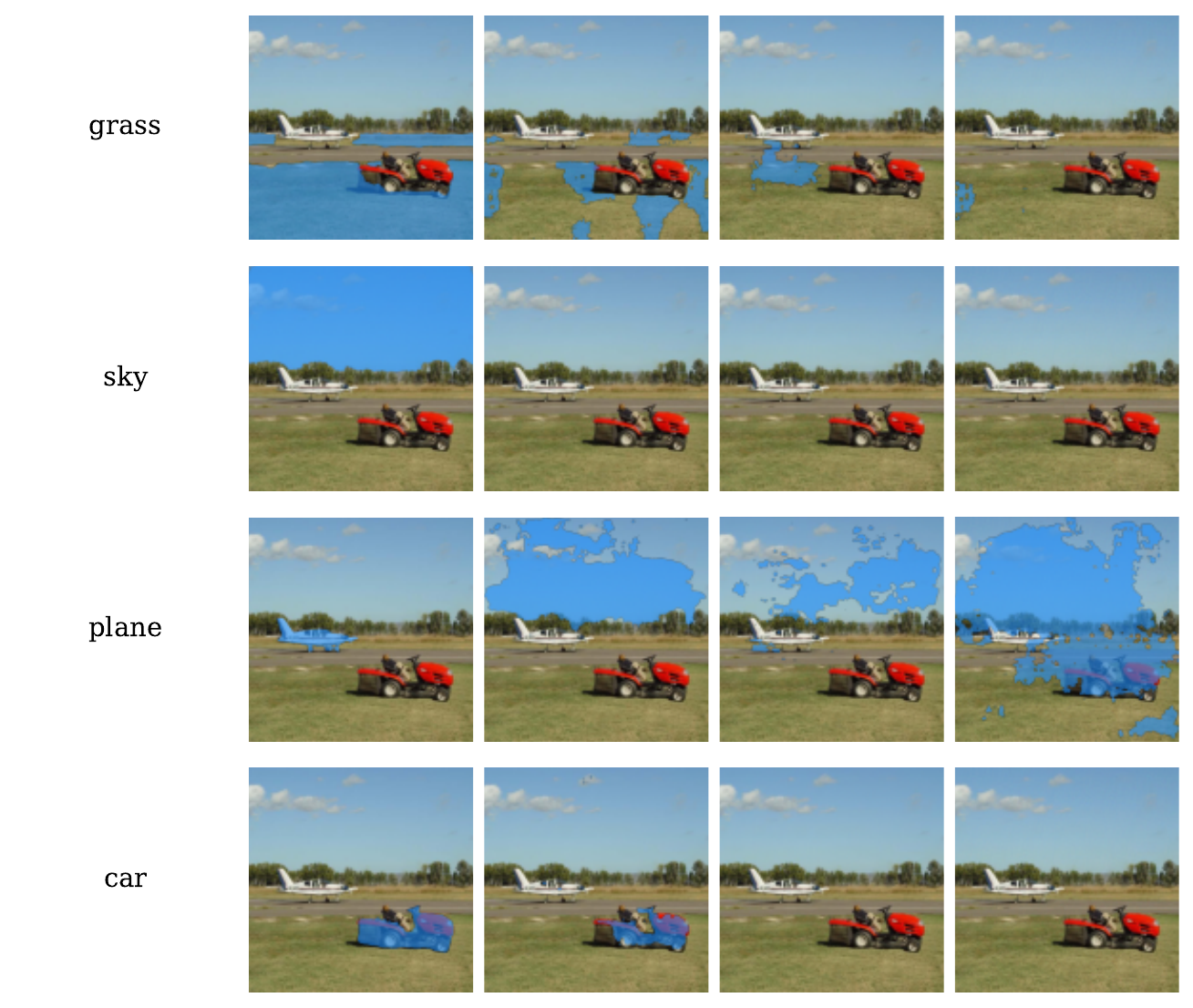}}
\\
\midrule
\multicolumn{5}{c|}{\includegraphics[align=c, width=\newl]{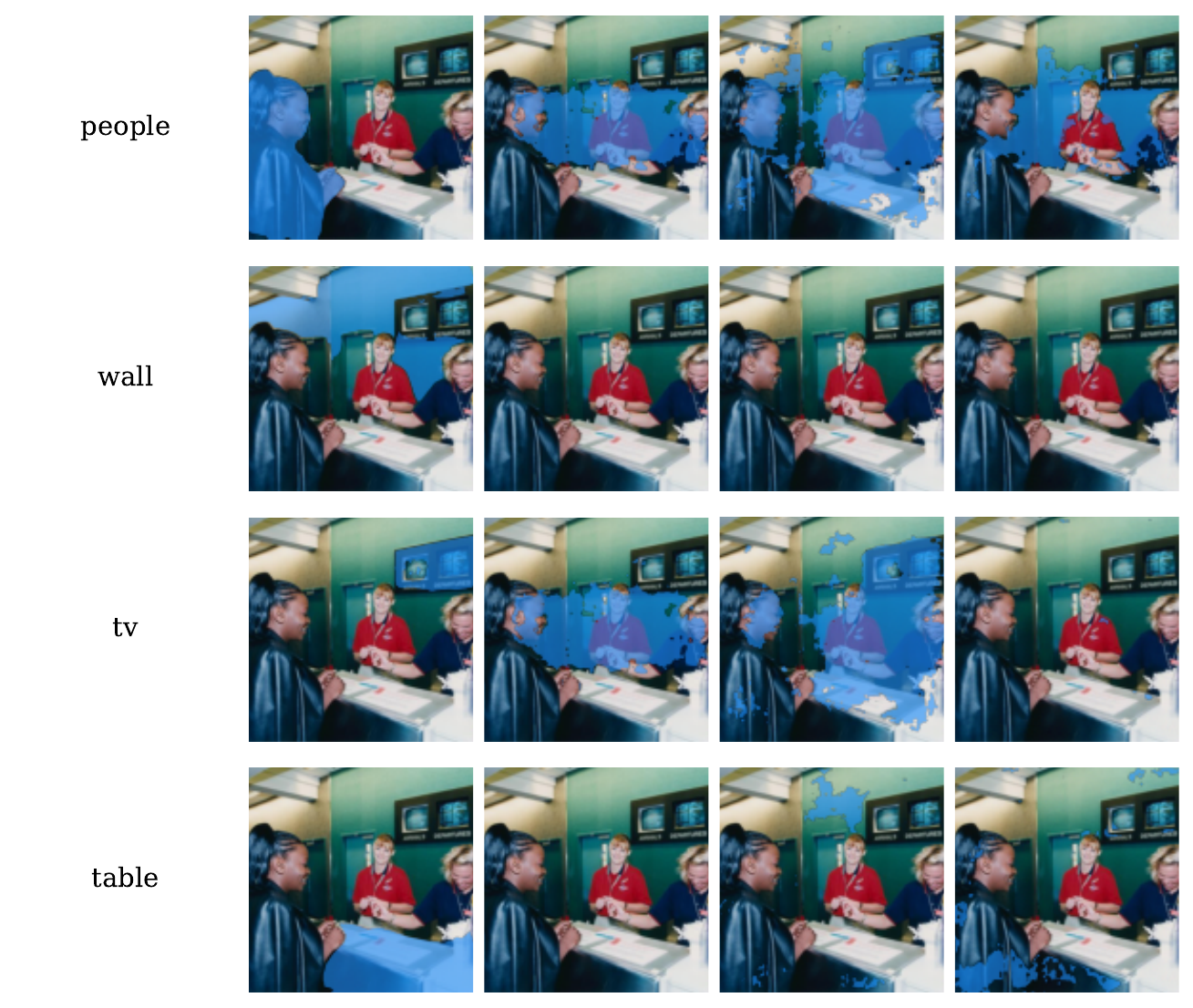}} & 
\multicolumn{5}{c}{\includegraphics[align=c, width=\newl]{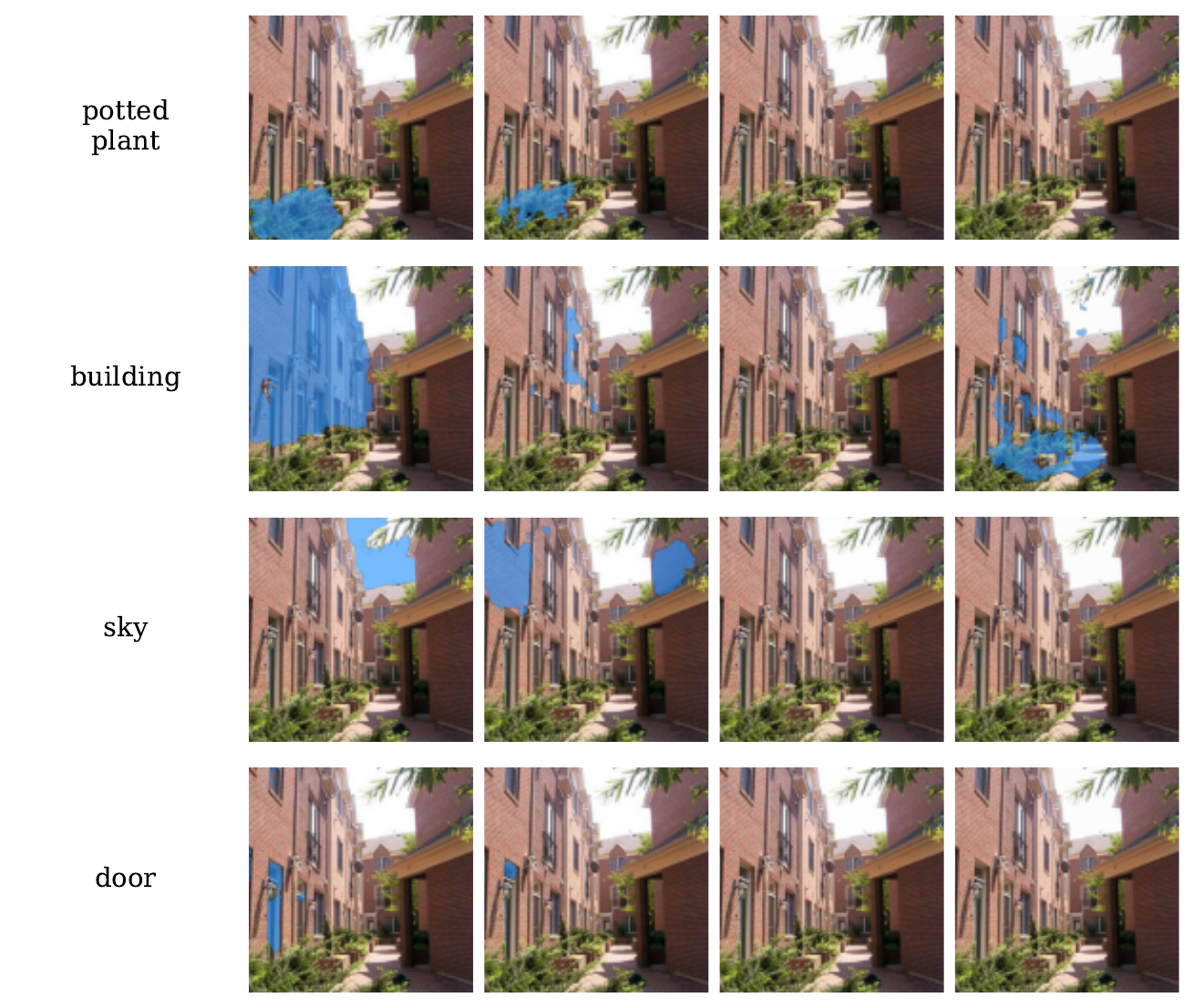}}

\end{tabular}
\caption{\textbf{Text prompts on \seem (Focal-L).} We show the predicted segmentation masks given several text prompts for both clean and adversarially perturbed (with perturbation size $\epsilon \in \{1/255, 2/255, 4/255\}$) images. In this case we use images from \ade. While the original masks can precisely identify the objects for each class, small perturbations significantly alter the predicted masks.
} \label{fig:seem_focall_text_prompts_ade}

\end{figure*}

\textbf{Text prompts.}
A prompt modality which is accepted by \seem is text, where one can use class names to segment the corresponding objects. We compare the segmentation masks associated by \seem with Focal-L backbone to different prompts for both clean (from \ade) and adversarial inputs in Fig.~\ref{fig:seem_focall_text_prompts_ade}.
While the model is able to precisely find the objects from the prompts for clean images, even small perturbations with $\epsilon=1/255$ are sufficient to notably degrade its performance for all prompts, independently from the size of the target object. We highlight that the same adversarially perturbed image is used with all prompts.

\section{Conclusion}

\textbf{Discussion.}
We have shown that it is possible to generate adversarial attacks on foundation models for various segmentation tasks in a prompt-agnostic fashion, even at low computational cost.
In fact, we demonstrated how attacking a single component of the complex architecture of such models may suffice to significantly degrade their performance.
This shows the vulnerability of these models, and potentially of the systems integrating them, to imperceptible perturbations.
 
Moreover, the existence of universal attacks which generalize to unseen images might be particularly interesting, and possibly beneficial.
In fact, we envision that such perturbations, especially if they could be found in a black-box setup (we provide an initial study in this direction in App.~\ref{sec:transfer_sam_to_seem_app}, where we analyze the transferability of universal attacks generated on \sam to \seem), might be used to prevent the automatic processing  of publicly shared images for segmentation tasks (and thus subsequent downstream tasks) by foundation models, and thus result in more privacy.

Further analyses of these aspects would allow the community to better understand the functioning of foundation models for segmentation and lead to a safer deployment of those in real world applications.
\\

\textbf{Limitations.}
While our main goal was to demonstrate the feasibility of prompt-agnostic attacks, and their effectiveness in a variety of segmentation tasks, we limited our empirical evaluation to \textit{1)} the most popular models (\sam and \seem), and a subset of all modalities they allow for, \textit{2)} the white-box scenario, with full access to the target models which might not be practical in some applications, \textit{3)} exploring only a few aspects of the attacks.
For example, we did not test all possible types of prompts like box plus points in \sam, as well as tuning relevant components of the attacks e.g. the loss used by APGD (targeted attacks might even be possible by designing specific objective functions).
\\

\textbf{Future work.}
Since our approach is simple but effective and we use a small computational budget for our experiments, we foresee that it can be an important starting point for future research to develop more sophisticated attacks, possibly in the black-box scenario.
Furthermore, one could aim at a single universal perturbation optimized to work across different segmentation models, including those not used when creating the attacks.
Finally, it will be an interesting (and challenging) direction to explore how to make foundation models for segmentation adversarially robust without losing their flexibility and generalization properties.

\section*{Acknowledgements}

The authors acknowledge support from the DFG Cluster of Excellence ``Machine Learning – New Perspectives for Science'', EXC 2064/1.

{\small
\bibliographystyle{abbrv}

}

\clearpage

\begin{appendices}

\section{Additional Results} \label{sec:additional_results_app}

\subsection{Universal attacks on \sam}

\begin{figure*} \centering \small
\newl=.25\columnwidth
\colwidth=.5\columnwidth
\tabcolsep=1.1pt 
\begin{tabular}{*{1}{C{\colwidth}} | *{1}{C{\colwidth}}
|| *{1}{C{\colwidth}} | *{1}{C{\colwidth}}
}
\multicolumn{2}{c||}{\textbf{seen images}} & \multicolumn{2}{c}{\textbf{unseen images}} \\[2mm]

original & perturbed ($\epsilon=8/255$) & original & perturbed ($\epsilon=8/255$) \\
\toprule

\multicolumn{2}{c||}{\includegraphics[align=c, width=4\newl]{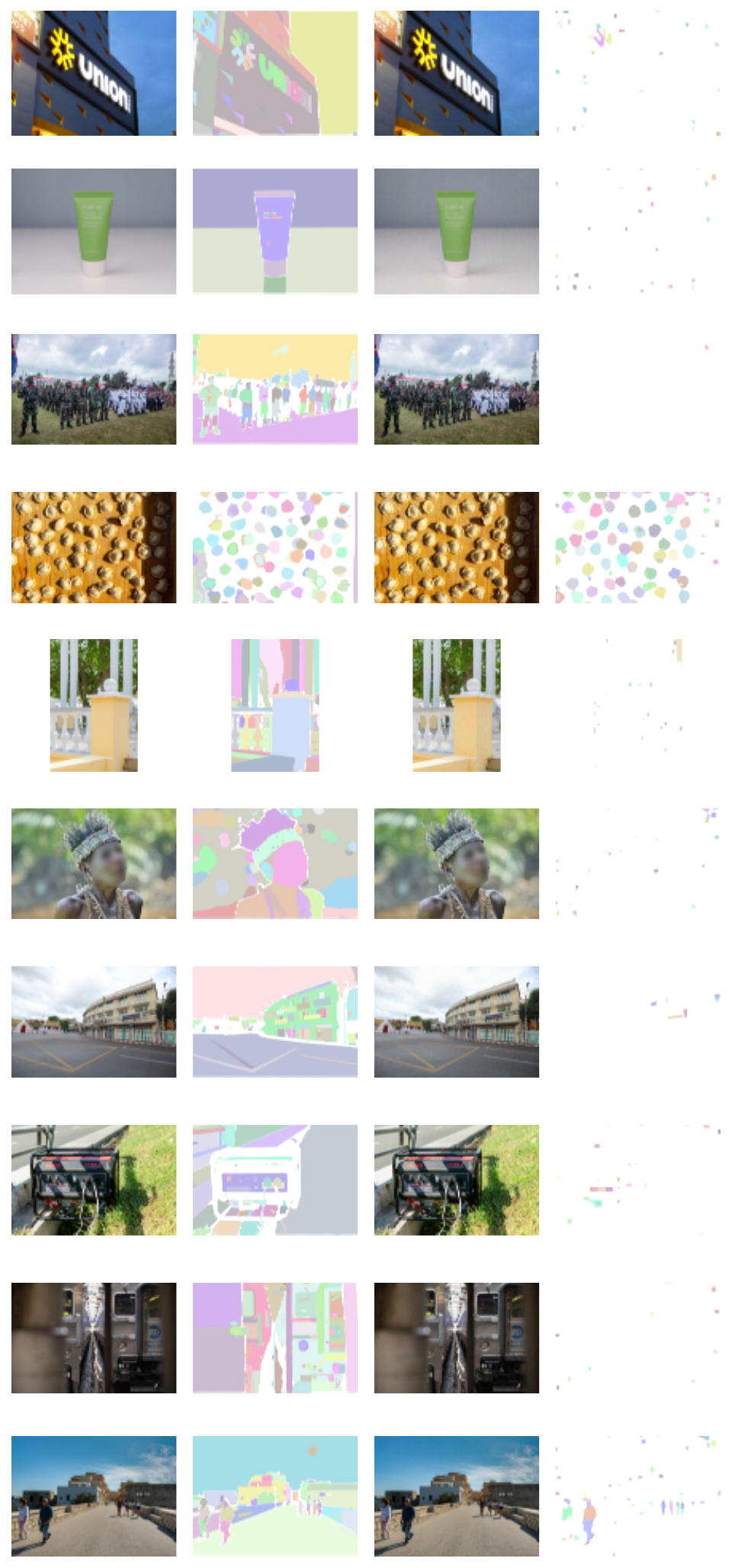}} & 
\multicolumn{2}{c}{\includegraphics[align=c, width=4\newl]{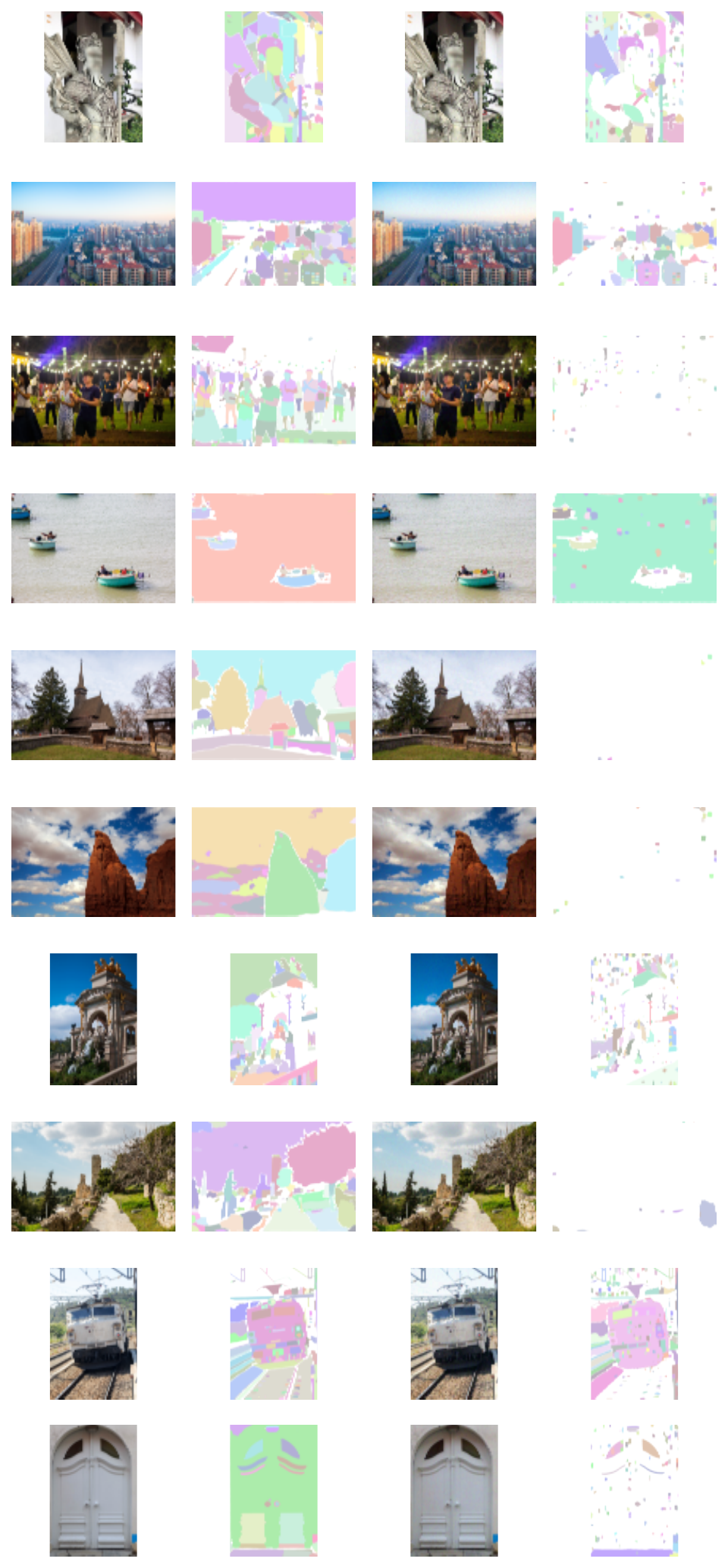}}

\end{tabular}
\caption{\textbf{Universal attacks on \sam.} We repeat the experiment shown in Fig.~\ref{fig:sam_univ_attacks} for a different set of training images.}\label{fig:sam_univ_attacks_seed-0}
\end{figure*}

\begin{figure*} \centering \small
\newl=.25\columnwidth
\colwidth=.5\columnwidth
\tabcolsep=1.1pt 
\begin{tabular}{*{1}{C{\colwidth}} | *{1}{C{\colwidth}}
|| *{1}{C{\colwidth}} | *{1}{C{\colwidth}}
}
\multicolumn{2}{c||}{\textbf{seen images}} & \multicolumn{2}{c}{\textbf{unseen images}} \\[2mm]

original & perturbed ($\epsilon=4/255$) & original & perturbed ($\epsilon=4/255$) \\
\toprule

\multicolumn{2}{c||}{\includegraphics[align=c, width=4\newl]{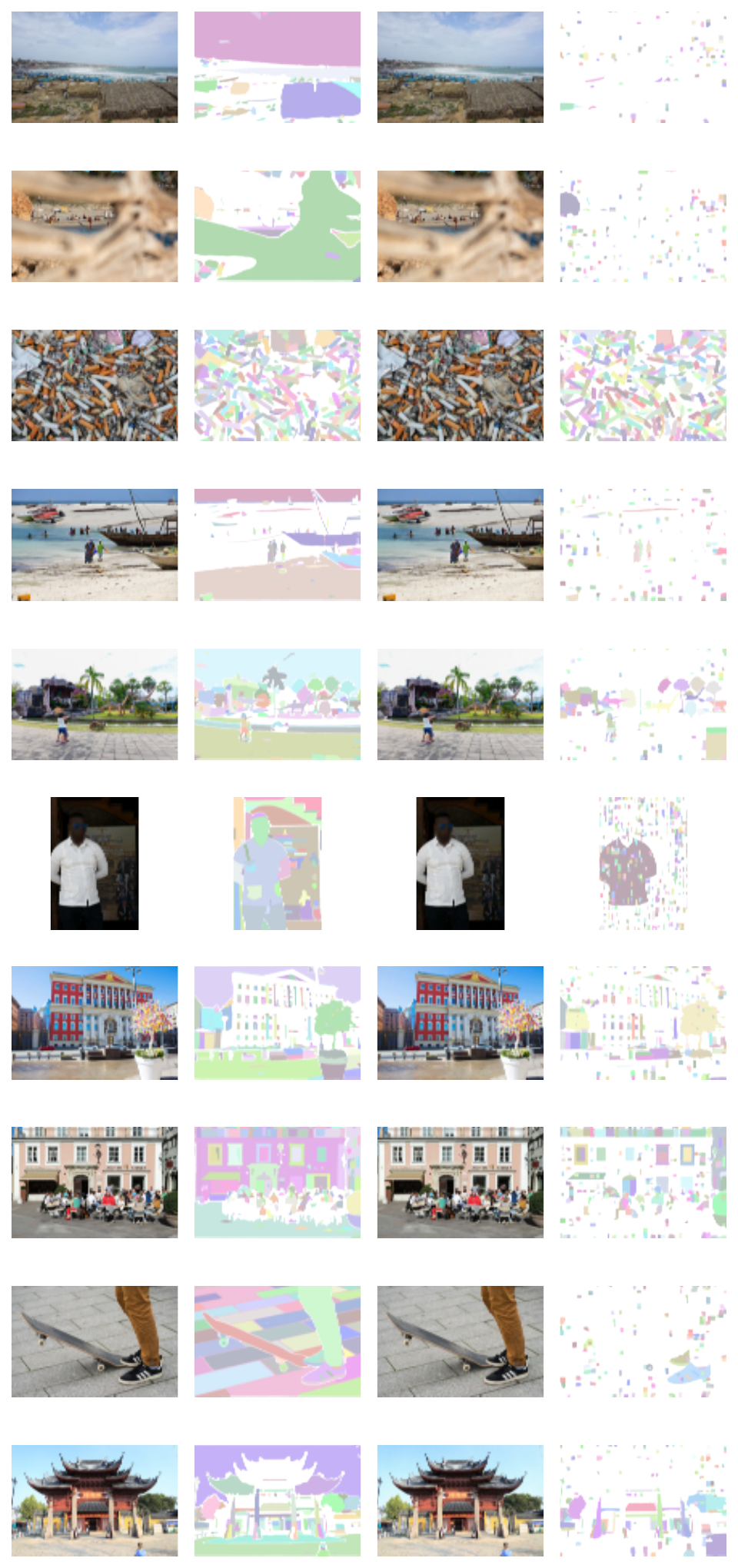}} & 
\multicolumn{2}{c}{\includegraphics[align=c, width=4\newl]{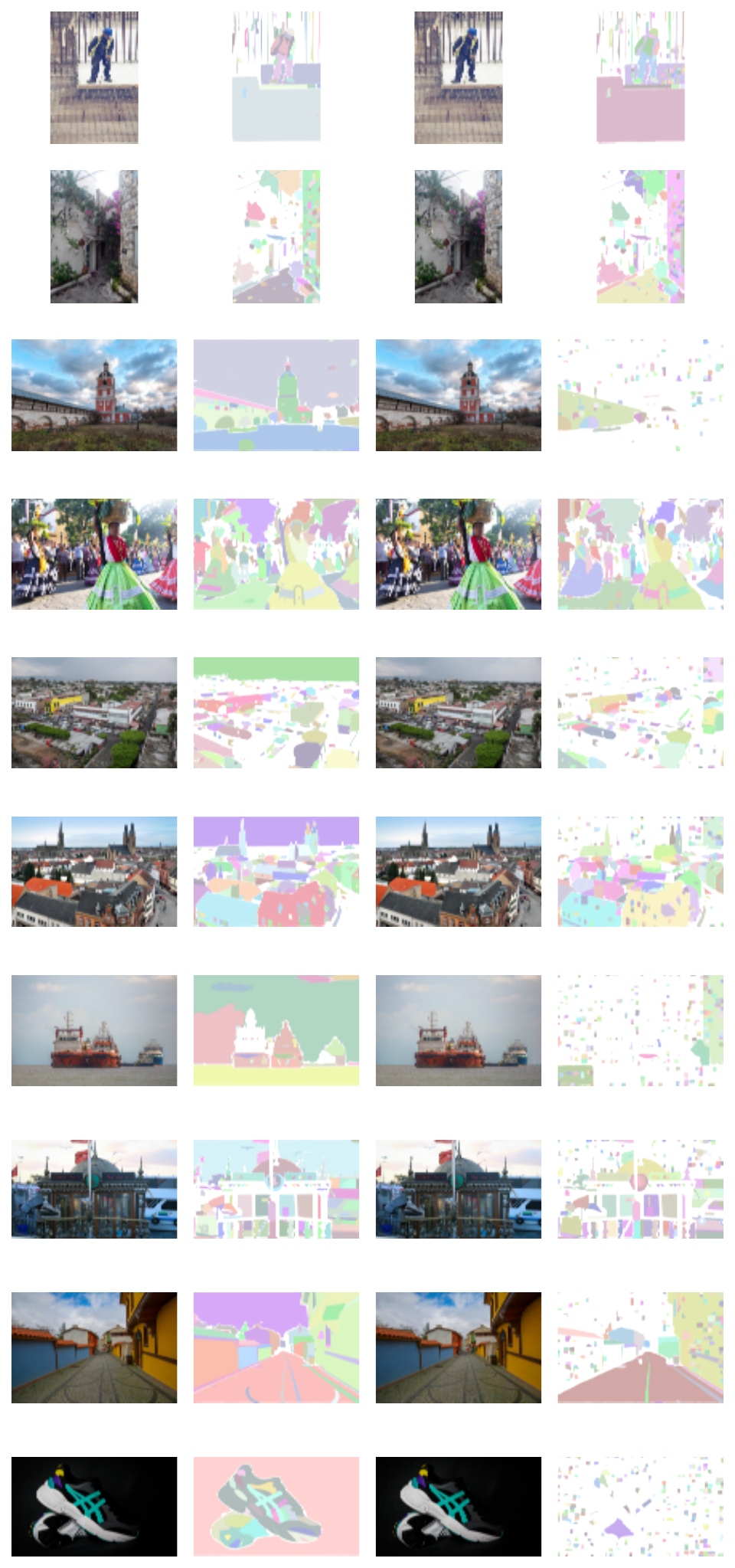}}

\end{tabular}
\caption{\textbf{Universal attacks on \sam with smaller radius.} We repeat the experiment shown in Fig.~\ref{fig:sam_univ_attacks} with bound of $4/255$ instead of $8/255$ on the $\ell_\infty$-norm of the universal perturbation.}\label{fig:sam_univ_attacks_eps_4}
\end{figure*}

In Fig.~\ref{fig:sam_univ_attacks_seed-0} we report the results of universal attacks generated as described in Sec.~\ref{sec:universal_attacks_sam} but using a different set of randomly sampled training images.
We see that even in this case we obtain similar results to those shown in Fig.~\ref{fig:sam_univ_attacks} above, with the universal attacks being able to generalize to unseen images.

Moreover, we repeat the experiment shown in Fig.~\ref{fig:sam_univ_attacks} except for the bound on the $\ell_\infty$-norm of the universal perturbation, which is reduced to $4/255$ (instead of $8/255$).
Fig.~\ref{fig:sam_univ_attacks_eps_4} shows that even with the smaller budget universal perturbations are able to noticeably reduce the quality of the predicted masks of the Segment Everything mode of \sam, although to a lower degree than with the standard radius $8/255$.

\subsection{White-box attacks on \seem}

\begin{figure*} \centering \small
\newl=1.8\columnwidth
\tabcolsep=1.1pt 
\begin{tabular}{*{5}{C{.155\columnwidth}} | *{5}{C{.205\columnwidth}}}

\multicolumn{5}{c|}{\textbf{\ade}} & \multicolumn{5}{c}{\textbf{\saoneb}} \\[2mm]

& original & $1/255$ & $2/255$ & $4/255$ 
& & original & $1/255$ & $2/255$ & $4/255$ \\
\toprule

\multicolumn{5}{c|}{\includegraphics[align=c, height=\newl]{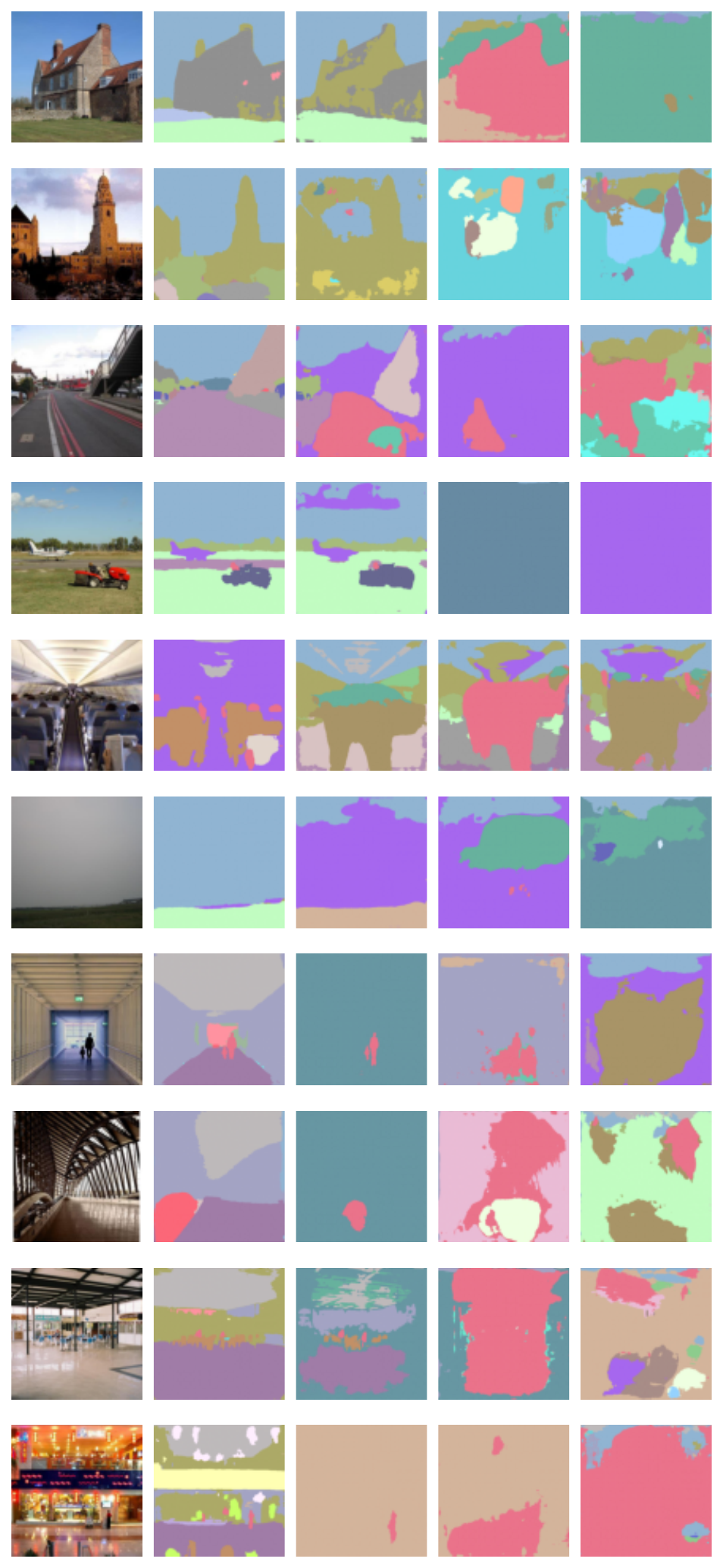}}&
\multicolumn{5}{c}{\includegraphics[align=c, height=\newl]{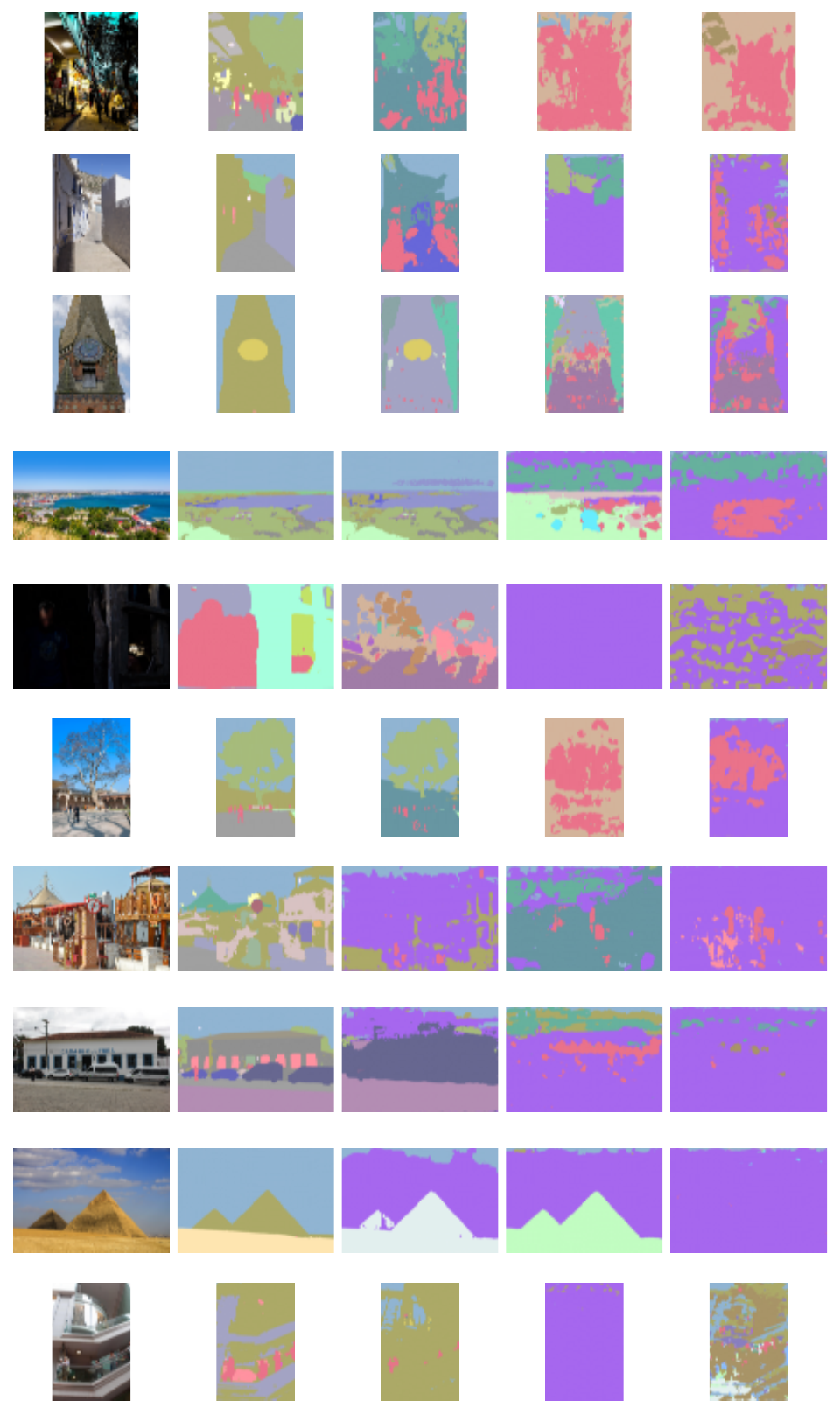}}

\end{tabular}
\caption{\textbf{Semantic segmentation with \seem (Focal-T).} For random images from the \ade (left) and \saoneb (right) dataset, we show the predicted segmentation masks (each color corresponds to a predicted class) for the original input and with adversarial perturbations of increasing strength. Small perturbations are sufficient to drastically change the semantic segmentation maps predicted by \seem with Focal-T backbone.} \label{fig:seem_focalt_semantic}
\end{figure*}

In Fig.~\ref{fig:seem_focalt_semantic} we show the effect of our attacks (with radii $\epsilon\in\{1/255, 2/255, 4/255\}$) on the semantic segmentation masks provided by \seem with the smaller Focal-T backbone, similarly to what done for the Focal-L backbone in Fig.~\ref{fig:seem_focall_semantic}.
Note that unlike what done for the larger backbone, the images from \saoneb are in this case used at the original resolution.
As observed above, the performance of \seem in semantic segmentation is significantly reduced on the perturbed images.

\subsection{Transferring universal attacks from \sam to \seem} \label{sec:transfer_sam_to_seem_app}

\begin{figure*} \centering \small
\newl=2\columnwidth
\colwidth=.28\columnwidth
\tabcolsep=1.1pt 
\begin{tabular}{*{3}{C{\colwidth}} |
*{2}{C{\colwidth}}}

& \multicolumn{2}{c|}{\textbf{\seem with Focal-T}} & \multicolumn{2}{c}{\textbf{\seem with Focal-L}} \\[2mm]

& original & \makecell{perturbed\\($\epsilon=8/255$)} & original & \makecell{perturbed\\ ($\epsilon=8/255$)} \\
\toprule

\multicolumn{3}{c|}{\includegraphics[align=c, height=\newl]{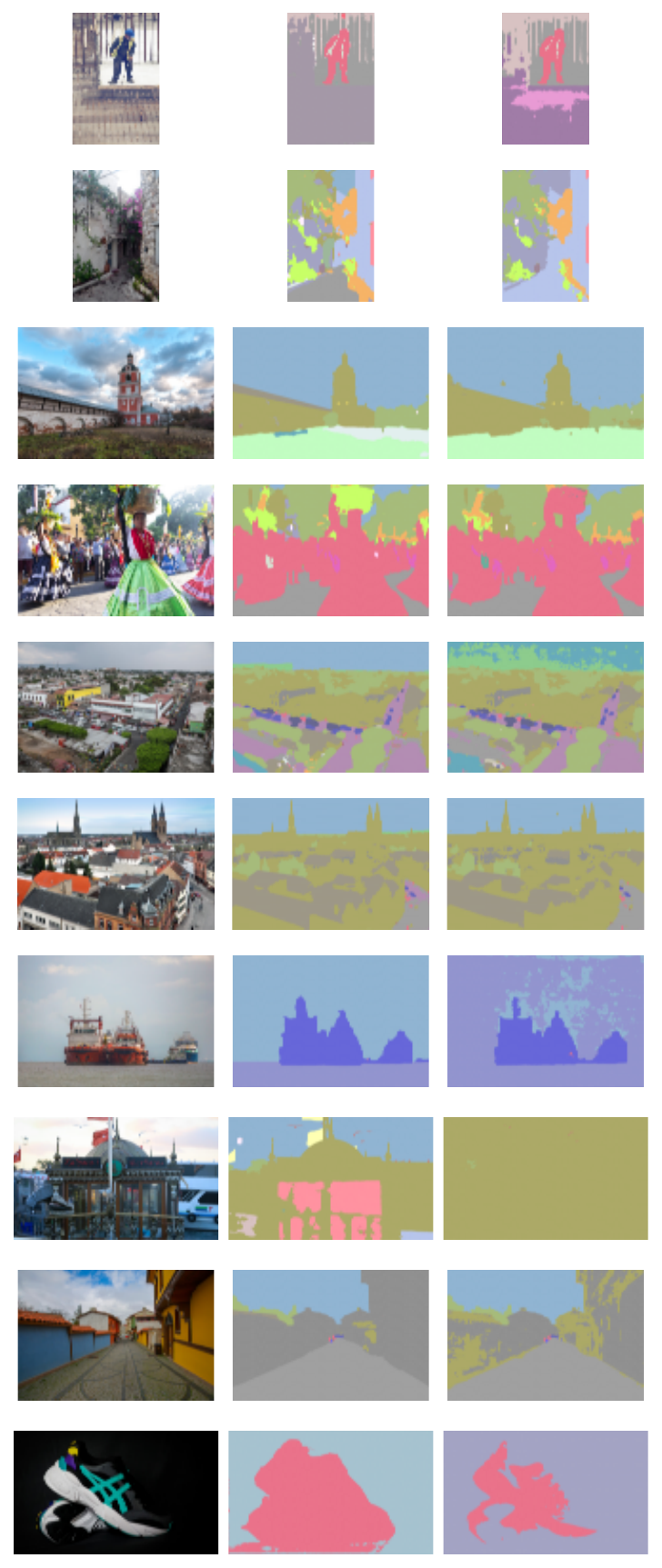}}&
\multicolumn{2}{c}{\includegraphics[align=c, height=\newl]{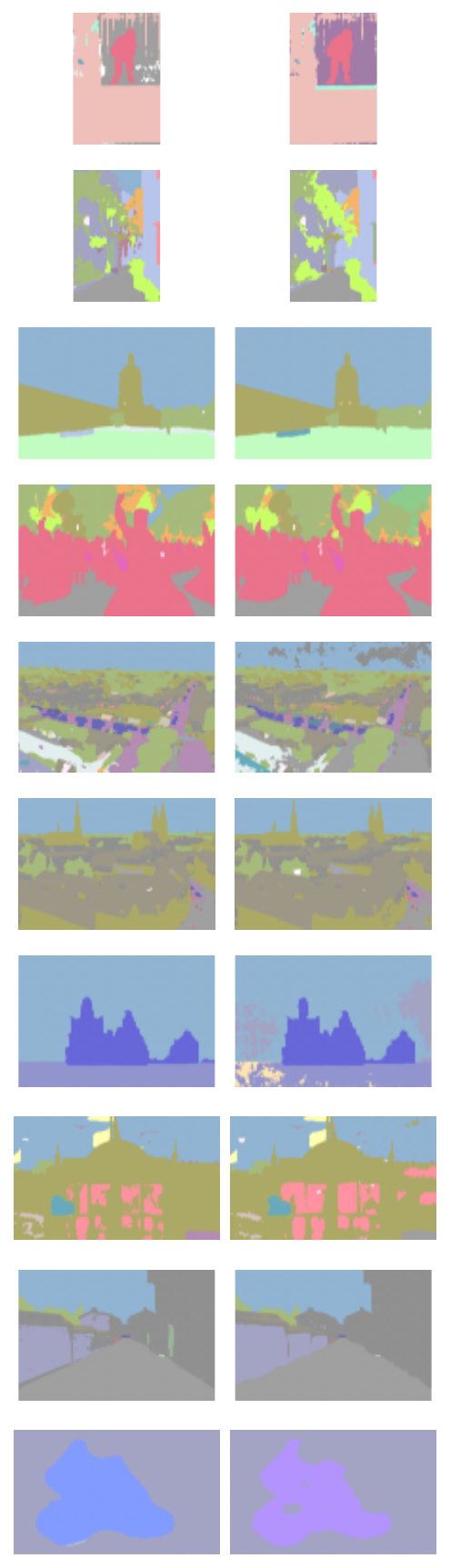}}

\end{tabular}
\caption{\textbf{Semantic segmentation by \seem when transferring universal attacks from \sam.} We test how well the universal attacks generated on \sam with $\epsilon=8/255$, shown in Fig.~\ref{fig:sam_univ_attacks}, transfer to \seem. For both Focal-T (left) and Focal-L (right) backbones, we show the predicted segmentation masks (each color corresponds to a predicted class) for the clean and perturbed images. While the predictions are mostly stable, especially for the larger Focal-L, the transferred universal perturbation introduce some mild degradation in the output of \seem.} \label{fig:seem_from_sam_semantic}
\end{figure*}

We further test the transferability of our universal attacks generated on \sam for \saoneb images (Sec.~\ref{sec:universal_attacks_sam}) to other models, i.e. \seem with different backbones.
In Fig.~\ref{fig:seem_from_sam_semantic} we show the semantic segmentation predicted by \seem on the images perturbed with the universal attack of size $\epsilon=8/255$ presented in Fig.~\ref{fig:sam_univ_attacks}. We observe that the predicted maps are stable for most input images, especially for the larger Focal-L backbone, but adding the universal perturbations can anyway lead to some mild quality degradation.
Transfer attacks across models are known to work better as the architecture of source and target models become more similar. In this case, \sam and \seem encoders have quite different features, then it is not surprising to observe the limited success of transfer attacks.
We leave to future work designing specific techniques (or adapting some from the rich literature on transfer attacks for image classifiers) to improve these results.

\end{appendices}

\end{document}